%% file: PaperForReview.tex
\documentclass[10pt,twocolumn,letterpaper]{article}

\usepackage{cvpr}              % To produce the CAMERA-READY version
\usepackage{graphicx}
\usepackage{amsmath}
\usepackage{amssymb}
\usepackage{booktabs}
\usepackage{multirow}
\usepackage{tabularx}
\usepackage{multirow}
\usepackage{makecell}

\usepackage[pagebackref,breaklinks,colorlinks]{hyperref}

\usepackage[capitalize]{cleveref}
\crefname{section}{Sec.}{Secs.}
\Crefname{section}{Section}{Sections}
\Crefname{table}{Table}{Tables}
\crefname{table}{Tab.}{Tabs.}

\def\cvprPaperID{11405} % *** Enter the CVPR Paper ID here
\def\confName{CVPR}
\def\confYear{2025}

\begin{document}

\title{OpenHumanVid: A Large-Scale High-Quality Dataset for Enhancing Human-Centric Video Generation}

\author{Hui Li\textsuperscript{1*},
        Mingwang Xu\textsuperscript{1*},
        Yun Zhan\textsuperscript{1},
        Shan Mu\textsuperscript{1},
        Jiaye Li\textsuperscript{1},
        Kaihui Cheng\textsuperscript{1},
        Yuxuan Chen\textsuperscript{1},
        Tan Chen\textsuperscript{1},\\
        Mao Ye\textsuperscript{4},
        Jingdong Wang\textsuperscript{2},
        Siyu Zhu\textsuperscript{1,3} \\
\textsuperscript{1}Fudan University, 
\textsuperscript{2}Baidu Inc, 
\textsuperscript{3}Shanghai Academy of AI for Science, \\
\textsuperscript{4}Shanghai Jiaotong University
\\
}
\maketitle

\renewcommand{\thefootnote}{\fnsymbol{footnote}}
\footnotetext[1]{ indicates equal contribution.}
\renewcommand{\thefootnote}{\arabic{footnote}}

%%%%%%%%% ABSTRACT
\begin{abstract}
Recent advancements in visual generation technologies have markedly increased the scale and availability of video datasets, which are crucial for training effective video generation models. 
However, a significant lack of high-quality, human-centric video datasets presents a challenge to progress in this field. 
To bridge this gap, we introduce \textbf{OpenHumanVid}, a large-scale and high-quality human-centric video dataset characterized by precise and detailed captions that encompass both human appearance and motion states, along with supplementary human motion conditions, including skeleton sequences and speech audio.
To validate the efficacy of this dataset and the associated training strategies, we propose an extension of existing classical diffusion transformer architectures and conduct further pretraining of our models on the proposed dataset. 
Our findings yield two critical insights: First, the incorporation of a large-scale, high-quality dataset substantially enhances evaluation metrics for generated human videos while preserving performance in general video generation tasks. 
Second, the effective alignment of text with human appearance, human motion, and facial motion is essential for producing high-quality video outputs.
Based on these insights and corresponding methodologies, the straightforward extended network trained on the proposed dataset demonstrates an obvious improvement in the generation of human-centric videos. Project
page: \href{https://fudan-generative-vision.github.io/OpenHumanVid}{https://fudan-generative-vision.github.io/OpenHumanVid}.
\vspace{-4mm}
\end{abstract}

%%%%%%%%% BODY TEXT
\input{sections/introduction}
\input{sections/related_work_arxiv}
\input{sections/dataset_arxiv}

\input{sections/network}
\input{sections/experiments_arxiv}

\input{sections/conclusion_arxiv}
%%%%%%%%% REFERENCES
{\small
\bibliographystyle{ieee_fullname}
\bibliography{egbib}
}

%\clearpage
% \input{sections/appendix}

\end{document}

\title{OpenHumanVid: A Large-Scale High-Quality Dataset for Enhancing Human-Centric Video Generation}

\author{First Author\\
Institution1\\
Institution1 address\\
{\tt\small firstauthor@i1.org}
\and
Second Author\\
Institution2\\
First line of institution2 address\\
{\tt\small secondauthor@i2.org}
}
\maketitle

% Please see Figure~\ref{fig:quality_filter} for the comparison of video quality before and after this filtering process.
\begin{figure*}[!t]
    \centering
    \includegraphics[width=1.0\linewidth]{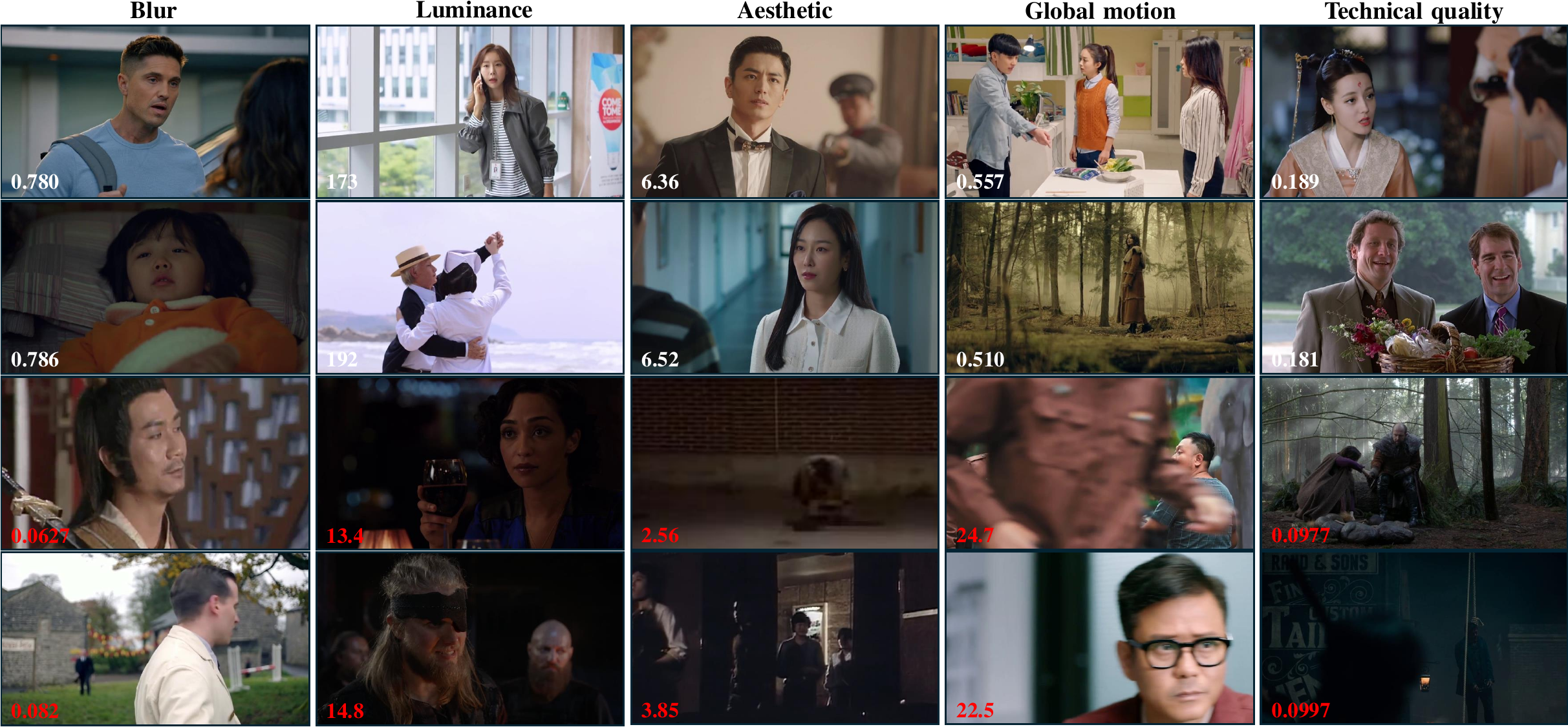}
    \caption{Videos we keep and deleted based on different quality filters}
    \label{fig:quality_filter}
\end{figure*}

%%%%%%%%% BODY TEXT
\section{Dataset}
\subsection{Video Quality Filter Details}
This study incorporates a comprehensive video quality filtering process grounded in several critical metrics: luminance, blurriness, aesthetic appeal, global and local motion, and overall technical quality. Each metric is evaluated against specific thresholds to retain only high-quality video content for further analysis.

\noindent\textbf{(1) Luminance.}  
Luminance is calculated using the formula: \(0.2126 \times R + 0.7152 \times G + 0.0722 \times B\), where \(R\), \(G\), and \(B\) denote the pixel values of the red, green, and blue channels, respectively. Videos are retained if their luminance values fall within the range of [10, 210].

\noindent\textbf{(2) Blur.}  
The level of blur in the video is assessed via the Cumulative Probability of Blur Detection algorithm \cite{narvekar2011no}, which analyzes edge feature distributions. Frames are extracted from the video, converted to grayscale, and the average, maximum, and minimum blur values are computed. Videos are filtered based on blur values within the range of [0.05, 1.0).

\noindent\textbf{(3) Aesthetic Quality.}  
Aesthetic appeal is measured using the CLIP-based Aesthetics Predictor \cite{aesthetic_predictor}. This method excludes videos with aesthetic scores below 4.8, thereby ensuring a satisfactory visual experience.

\noindent\textbf{(4) Global and Local Motion.}  
Motion quality is evaluated using optical flow analysis, specifically through the UniMatch algorithm \cite{unimatch}, which provides scores for both local and global motion. The filtering criteria for these scores are set within the range of [0.5, 20].

\noindent\textbf{(5) Technical Quality.}  
Lastly, the overall technical quality of the video is assessed using the DOVER model \cite{wu2023exploring}, filtering out videos with scores below 0.09.

\subsection{Human Motion Generation Details}

\begin{figure*}[!htbp]
    \centering
    \includegraphics[width=1.0\linewidth]{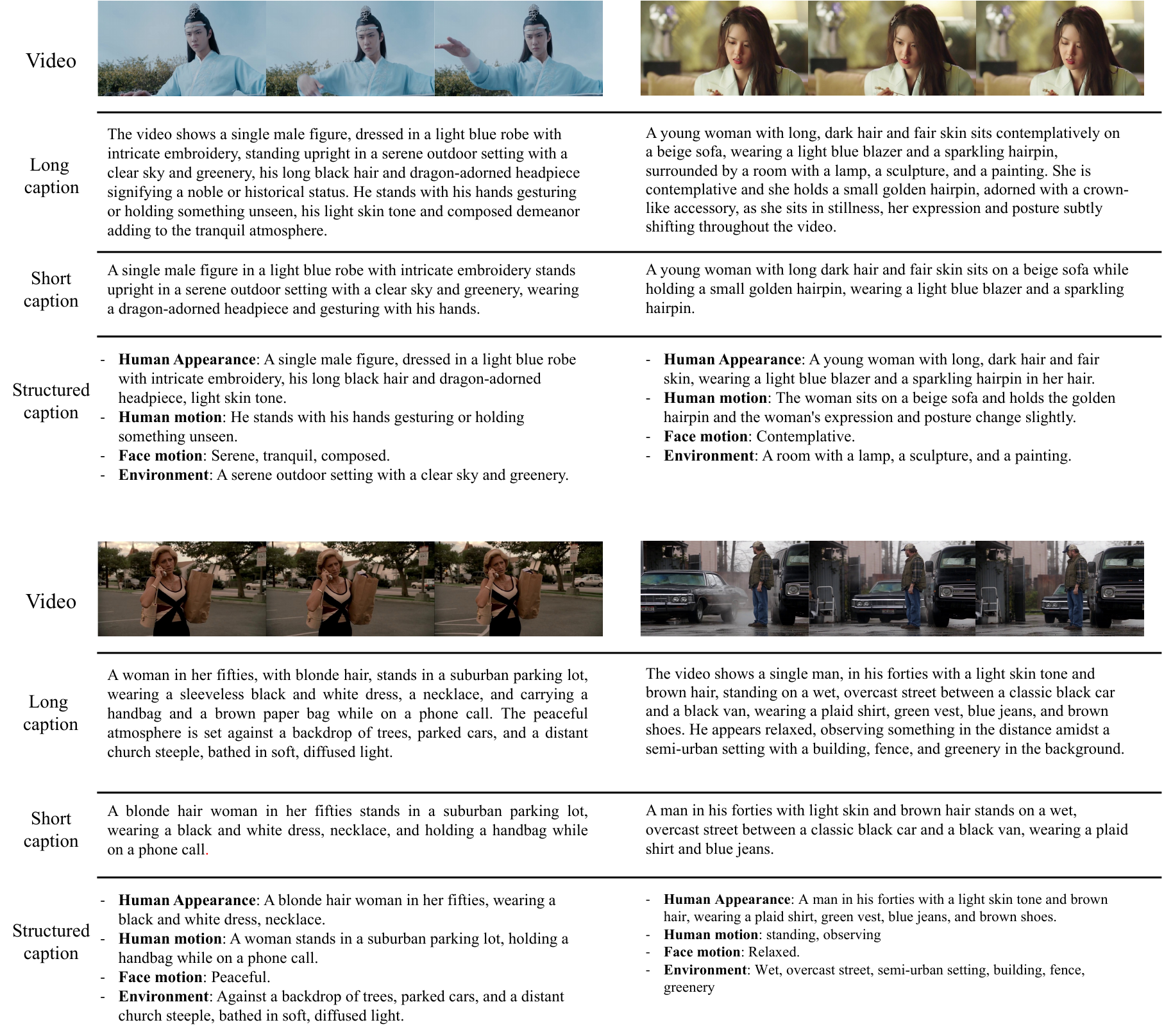} 
    \caption{The illustration of textual captions and corresponding types (long, short, and structured captions). }
    \label{fig:supp_caption} 
\end{figure*}

\noindent\textbf{Textual Condition.}  
Figure~\ref{fig:supp_caption} illustrates examples of various caption formats. We provide final captions in short, long, and structured formats to capture different contextual details.

\noindent\textbf{Skeleton Sequence.}  
Figure~\ref{fig:supp_skeleton} presents examples of skeleton annotations extracted from videos using DWpose \cite{yang2023effective}. These annotations offer a structured representation of joint positions that inform the movement of the generated figures.

\noindent\textbf{Speech Audio.}  
Figure~\ref{fig:supp_speech} shows examples of speech audio from the videos. We utilize SyncNet \cite{raina2022syncnet} to evaluate the synchronization between the subject's speech audio and their corresponding lip movements.

\begin{figure*}[!htbp]
    \centering
    \includegraphics[width=1.0\linewidth]{figs/supp/supp_skeleton.pdf} 
    \caption{The illustration of human skeleton sequences with respect to given videos.}
    \label{fig:supp_skeleton} 
\end{figure*}

\begin{figure*}[!htbp]
    \centering
    \includegraphics[width=1.0\linewidth]{figs/supp/supp_speech.pdf} 
    \caption{The illustration of speech audio with respect to given videos. Left: video screenshot; right: speech script.}
    \label{fig:supp_speech} 
\end{figure*}

\section{Experiments}

\subsection{Training on OpenHumanVid}
We utilize a large-scale, high-quality human video dataset, implementing the training data strategies outlined in the paper. 
These strategies include a higher video sampling rate, human appearance text-video filter, human motion text-video filter, and face motion text-video filter. 
As a result, the extended model demonstrates a significant improvement over the baseline following further pretraining.

\begin{figure*}
    \centering
    \includegraphics[width=1\linewidth]{figs/consistency2.pdf}
    \caption{Face and body consistency comparison between baseline and ours. }
    \label{fig:consistency}
\end{figure*}

\begin{figure*}
    \centering
    \includegraphics[width=1\linewidth]{figs/semantics2.pdf}
    \caption{Face and body semantics comparison between baseline and ours.}
    \label{fig:semantics}
\end{figure*}

\subsection{Limitations and Future Works}
Despite the advancements presented by OpenHumanVid, several limitations warrant consideration. 
Firstly, the reliance on existing multimodal models for caption generation may restrict the diversity and richness of the descriptions, potentially impacting the comprehensiveness of video-text alignment. 
Additionally, while the dataset incorporates various human motion conditions, the inherent variability of human actions and expressions across different cultures and contexts may not be fully captured. 
Future work could focus on enhancing the caption generation process through the development of more sophisticated models with improved instruction-following capabilities. 
Furthermore, expanding the dataset to include a wider array of human identities and motion scenarios, along with more granular textual prompts, could facilitate the generation of videos that better reflect the complexities of human behavior and interaction in diverse contexts.

\section{Safety Considerations}
The development of OpenHumanVid introduces several social risks, particularly concerning privacy, representation, and the potential for misuse. The dataset's inclusion of diverse human identities and movements necessitates rigorous ethical considerations to prevent the reinforcement of stereotypes or biases in generated videos. Additionally, there is a risk of inappropriate usage in contexts such as deepfakes or unauthorized portrayals of individuals. To mitigate these issues, we implement strict data governance protocols, ensuring that all content adheres to ethical standards and is used responsibly. Furthermore, we advocate for transparency in the dataset's application, emphasizing the necessity of guidelines for researchers and practitioners to promote responsible use while safeguarding individual rights and societal norms.

%%%%%%%%% REFERENCES
{\small
\bibliographystyle{ieee_fullname}
\bibliography{egbib}
}

%% file: sections/introduction.tex
\begin{figure*}[htbp]
    \centering
    \includegraphics[width=1.0\textwidth]{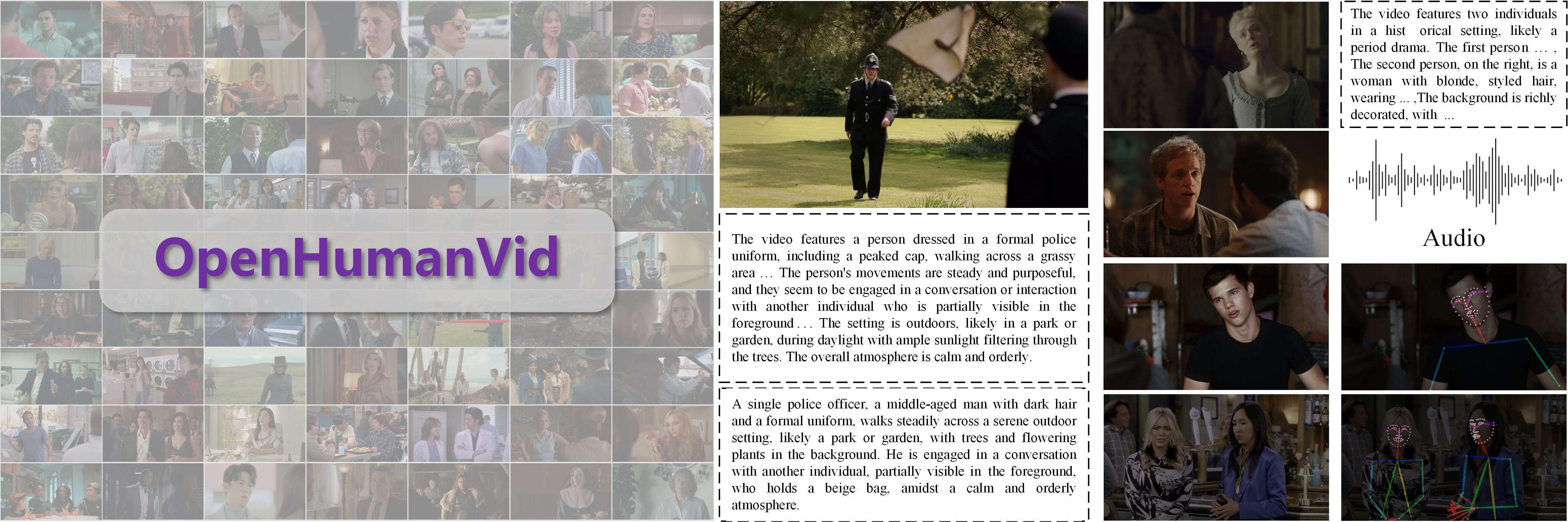} 
    \caption{Overview of the proposed OpenHumanVid dataset. The dataset comprises 52.3 million human video clips, totaling 70.6K hours of content. After applying video quality and human quality filters, the refined dataset includes 13.2 million high-quality human video clips. Each video is accompanied by three types of textual prompts: short, long, and structured. Additionally, each video contains human skeleton sequences and corresponding speech audio.}
    \label{fig:teaser}
    \vspace{-2mm}
\end{figure*}

\section{Introduction}
The emergence of general-purpose video generation models has catalyzed significant advancements in the field, particularly with the introduction of notable frameworks such as Stable Diffusion~\cite{rombach2022high}, Sora~\cite{liu2024sora}, and MovieGen~\cite{polyak2024movie}. 
Stable Diffusion employs a U-Net-based diffusion model~\cite{weng2021convolutional} within the latent space of a pre-trained Variational Autoencoder (VAE)~\cite{kingma2013auto}, thereby facilitating high-resolution visual generation while optimizing computational efficiency. 
In contrast, the Diffusion Transformer (DiT) network~\cite{peebles2023scalable} leverages a transformer architecture~\cite{vaswani2017attention} for denoising within the latent space, demonstrating superior scalability and enhanced visual quality.
Recently, video generation networks~\cite{polyak2024movie,kondratyuk2023videopoet,villegas2022phenaki} that utilize auto-regressive techniques for next-token prediction have exhibited considerable promise, particularly regarding scalability for long video generation and the achievement of high resolutions. 
A critical factor underpinning these advancements is the availability of high-quality, large-scale datasets, which are essential for the effective pretraining and fine-tuning of these models.

Nevertheless, existing video generation methods, especially those focused on generating videos featuring human subjects, encounter several notable limitations. 
These limitations can be categorized as follows:
(1)~\textbf{Appearance consistency}: Significant challenges persist concerning the appearance of human identities, including issues related to human body deformation and inconsistencies in identity representation.
(2)~\textbf{Motion alignment}: The alignment of human motion and motion control information remains problematic, resulting in unnatural human poses and expressions. 
Furthermore, there is often a temporal misalignment between lip movements, facial expressions, gesture motions, and the corresponding speech audio.
(3)~\textbf{Textual prompt support}: 
While contemporary general-purpose video generation models demonstrate robust support for textual prompts, they exhibit inadequate performance concerning fine-grained descriptions related to human appearance, motion, and expressions. 
Additionally, there is a lack of sufficient support for motion control signals, such as speech audio and human skeletal sequences.

Existing large-scale video datasets suitable for pretraining tasks in video generation, such as WebVid-10M~\cite{bain2021frozen}, Panda-70M~\cite{chen2024panda} and Open-Vid~\cite{nan2024openvid}, are characterized by their vast scale and diverse categories. 
However, these datasets exhibit a relative scarcity of data pertaining to human categories, resulting in insufficient diversity in human identities and motions, as well as inadequate textual prompts and motion conditions such as skeletal and speech audio. 
Concurrently, various human-centric datasets~\cite{soomro2012ucf101,shahroudy2016ntu,sadoughi2015msp,caba2015activitynet,chang2023magicdance,wang2024disco,jiang2023text2performer,ju2023human,camgoz2018neural} containing different identities, actions, and scenes have been applied. 
However, these datasets primarily target specific tasks such as human action, dance, fashion, and body language, rendering them unsuitable for further pretraining or fine-tuning general video generative models in the human domain.
To address these limitations, we introduce OpenHumanVid, which offers three key advantages:
(1)~\textbf{Large scale and high resolution}: OpenHumanVid is a high-resolution dataset (720P or 1080P) derived from reputable video sources, including films, television series, and documentaries.
This dataset encompasses a wide range of tasks and features rich identity appearances, varying human scales, human motion, expressions, and environmental information.
(2)~\textbf{Various human motion conditions}: We provide diverse motion-driven conditions, including textual prompts, skeletal data, and speech audio. Our prompts support short, long, and structured formats, encompassing identity appearance, motion, emotion, and background.
(3)~\textbf{Optimized text-video alignment}: We achieve a well-defined alignment between textual prompts and video visual information, including human appearance, facial motion, and human body motion.

To validate the effectiveness of our dataset, we conducted an analysis based on the prestigious diffusion transformer network~\cite{yang2024cogvideox}, examining improvements in human appearance consistency, motion smoothness, and motion-text alignment following further pretraining with our data. 
Based on these findings, we developed an enhanced version of the baseline model, demonstrating that even straightforward extensions of the network architecture can achieve notable improvements in the VBench~\cite{huang2024vbench} task after further pretraining on our proposed dataset. 
We intend to contribute the relevant datasets and code to the research community to facilitate further advancements in this domain.

%% file: sections/related_work_arxiv.tex
% \begin{figure*}[!t]  
%     \centering  
%     \includegraphics[width=17cm,height=3.13cm]{figs/source_data_statistics.png} % Replace 'example-image' with your image file name  
%     \caption{The statistical analysis of the source data.
%     This analysis elucidates the distribution of various video types, including films, television shows, and documentaries with various time durations.
%     Additionally, we examine the categorization of these videos, their resolutions, and the keywords associated with textual captions.
%     \textcolor{red}{@Liwei}}
%     \label{fig:blank_image}  
% \end{figure*} 

\begin{figure*}[!t]
    \centering
    \begin{subfigure}{0.23\linewidth}
        \centering
        \includegraphics[width=1.0\textwidth]{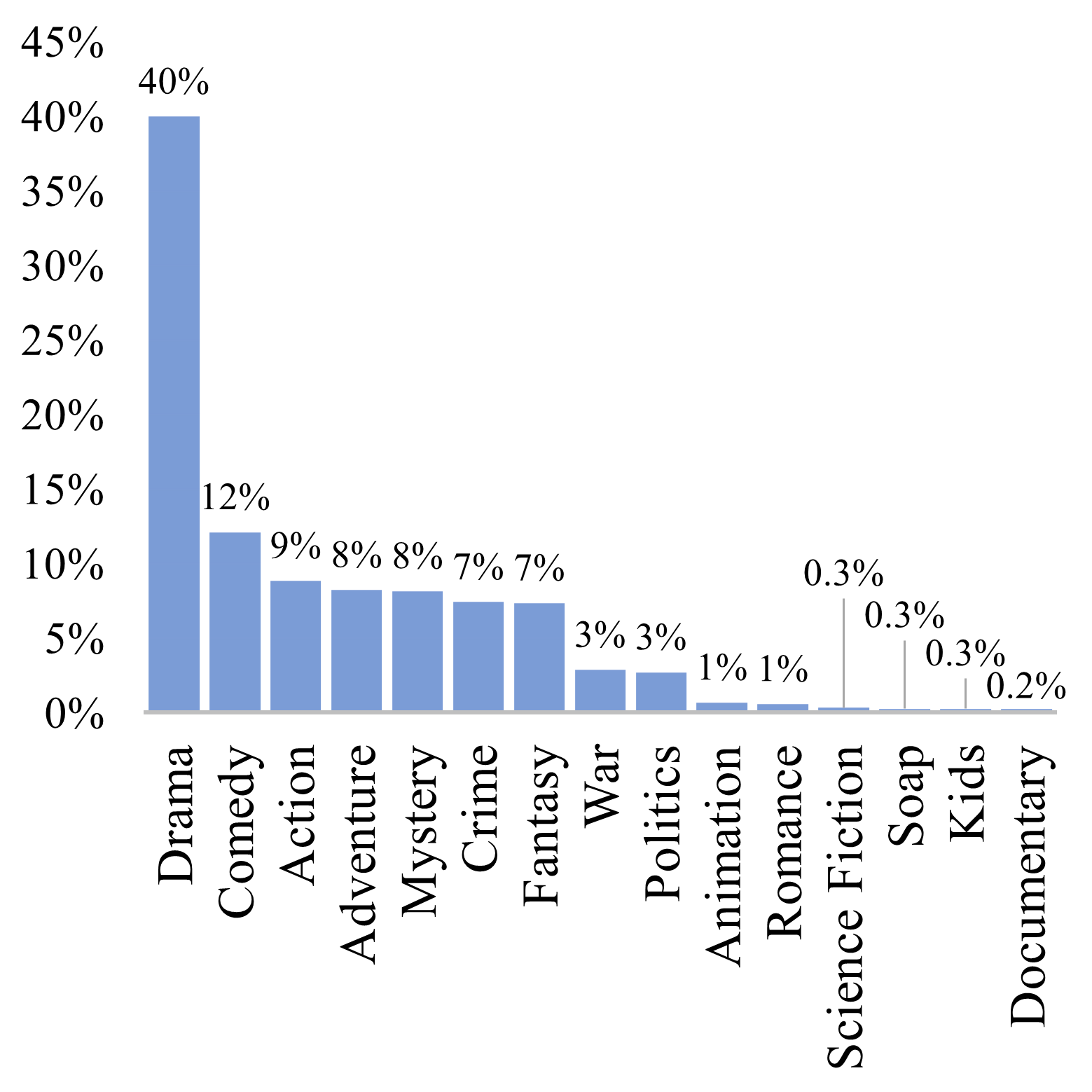} 
        \caption{Video categories}  
        \label{fig:source_data_category}  
    \end{subfigure}%
    \hfill
    \begin{subfigure}{0.23\linewidth}
        \centering
        \includegraphics[width=1.0\textwidth]{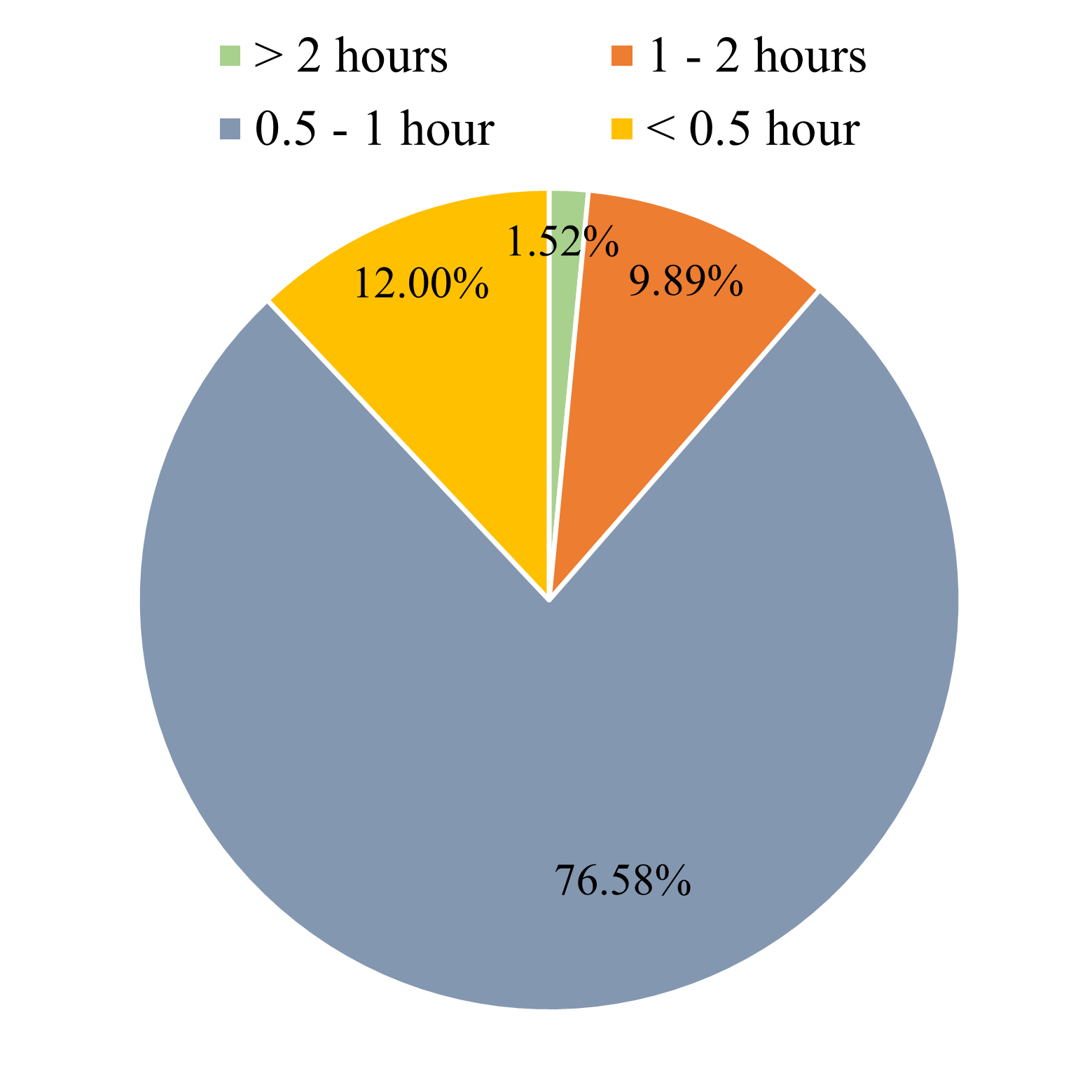}  
        \caption{Video duration}  
        \label{fig:source_data_duration}  
    \end{subfigure}%
    \hfill
    \begin{subfigure}{0.23\linewidth}
        \centering
        \includegraphics[width=1.0\textwidth]{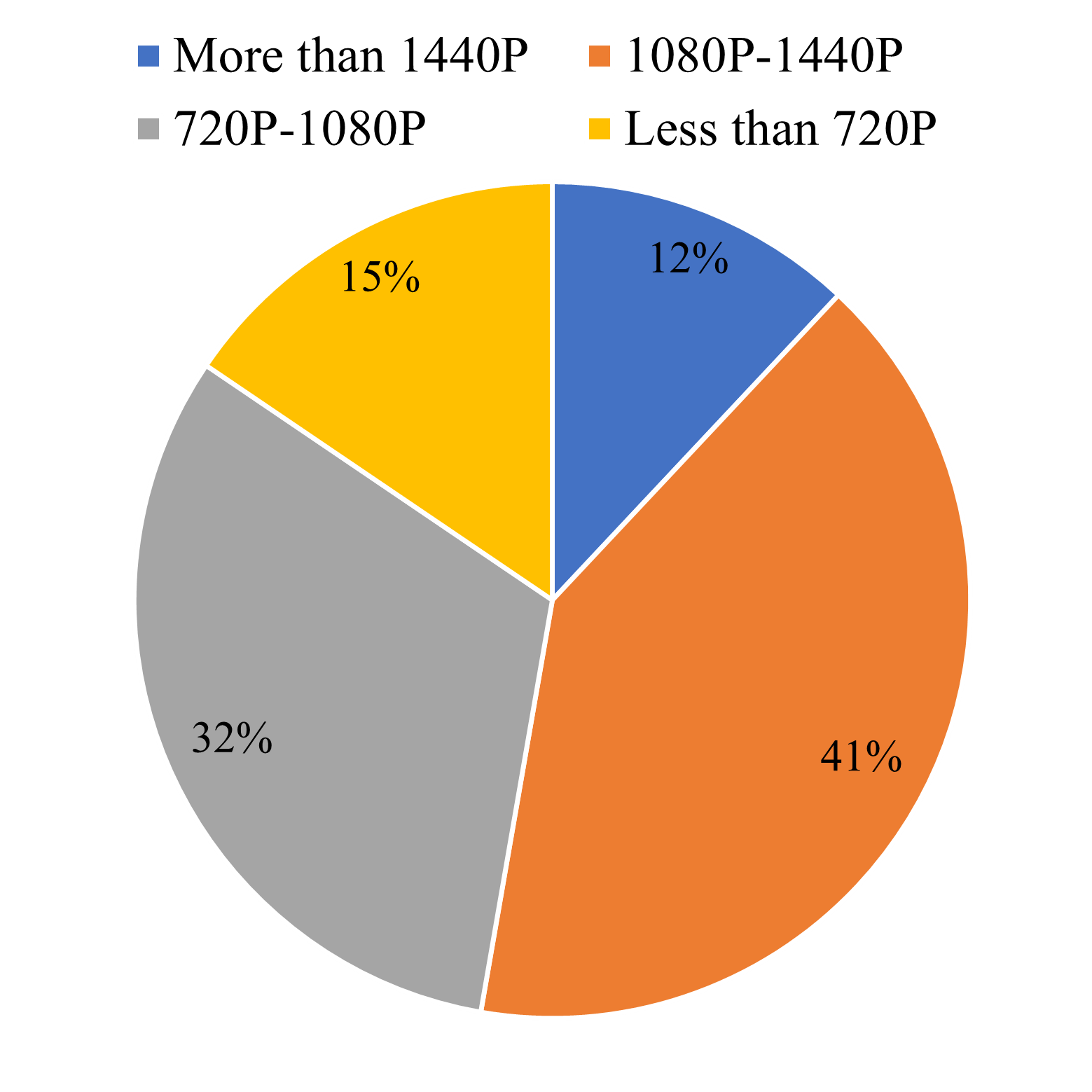}  
        \caption{Video resolution}  
        \label{fig:source_data_resolution}  
    \end{subfigure}  
    \begin{subfigure}{0.23\linewidth}
        \centering
        \includegraphics[width=1.0\textwidth]{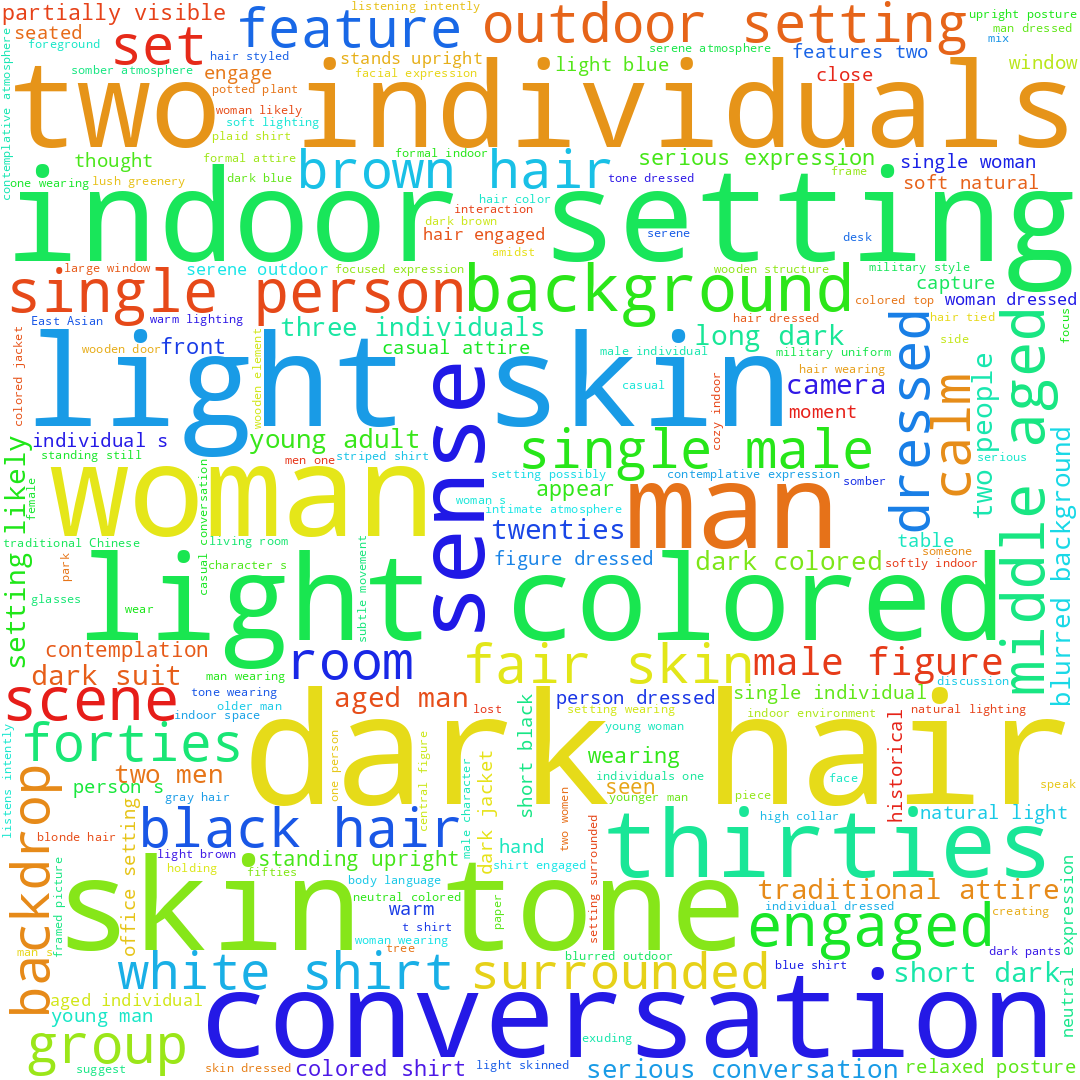}  
        \caption{Caption keywords}  
        \label{fig:source_data_keywords}  
    \end{subfigure}%
    % \vspace{-2mm}
    \caption{The statistical analysis of the source data.
    This analysis elucidates the distribution of various video types, including films, television shows, and documentaries of various time durations.
    Additionally, we examine the categorization of these videos, their resolutions, and the keywords associated with textual captions.}
    \label{fig:source_data}
    % \vspace{-2mm}
\end{figure*}

\begin{table*}[!t]
\centering
\resizebox{0.95\textwidth}{!}{
\begin{tabular}{@{}ccccccccc@{}}
\toprule
 Dataset name & Year&Domain & \# Videos &Total length (hours)&Caption type&Motion type& Resolution \\ 
\midrule
 WebVid-10M\cite{bain2021frozen}&2021&Open&10M&52K&Short&Text&360P\\
  Panda-70M\cite{chen2024panda}&2024&Open&70M&167K&Short&Text&720P\\
  OpenVid-1M\cite{nan2024openvid}&2024&Open&1M&2K&Long&Text&512P\\
  Koala-36M\cite{wang2024koala}&2024&Open&36M&172K&Long&Text&720P\\
  \midrule
  UCF-101\cite{soomro2012ucf101}& 2012&Human action & 13.3K & 26.7&Short&Text&240P\\
  NTU RGB+D\cite{shahroudy2016ntu}&2014&Human action&114K&37&-&3D pose, depth&1080P\\
  MSP-Avatar\cite{sadoughi2015msp}&2015&Human action&74&3&-&Speech audio,  landmark, pose&1080P\\
  ActivityNet\cite{caba2015activitynet}  & 2017&Human action & 100K & 849 & Short& Text& - \\
  TikTok-v4 \cite{chang2023magicdance} & 2023 &Human dance& 350 & 1& -&Skeleton & -  \\
  \midrule
Ours&2024&Human&52.3M&70.6K&Short, long, structured&Text, skeleton pose, speech audio&720P\\
Ours (filtered)&2024&Human&13.2M&16.7K&Short, long, structured&Text, skeleton pose, speech audio&720P\\
\bottomrule
\end{tabular}
}
% \vspace{-2mm}
\caption{The comparative analysis of our dataset against previous general and human video datasets. 
We enhance the textual captions by incorporating short, long, and structured formats that reflect human characteristics. 
Additionally, we integrate skeleton sequences derived from DWPose~\cite{yang2023effective} and corresponding speech audio filtered through SyncNet~\cite{raina2022syncnet} to enrich the dataset with contextual human motion data.}
\vspace{-1mm}
\label{tab:datasets}
\end{table*}

\section{Related Work}
\noindent\textbf{Diffusion Video Generation.}
Video generation primarily utilizes two methodologies: diffusion-based and autoregressive methods. Diffusion-based methods can be further categorized into U-Net-based and Diffusion Transformer (DiT)-based architectures, depending on the design of the denoising network.
U-Net-based diffusion models~\cite{ho2022video, ho2022imagen, blattmann2023align, xu2024hallo, cui2024hallo2, zhu2024champ, zhang2024tora} have seen significant advancements through seminal works such as Video Diffusion Models (VDMs)~\cite{ho2022video}, Imagen Video~\cite{ho2022imagen}, and Align Your Latents~\cite{blattmann2023align}. These approaches extend traditional image diffusion architectures to accommodate temporal data by integrating temporal dimensions into the latent space or by jointly training on image and video datasets. Recent open-source models, including Stable Video Diffusion~\cite{blattmann2023stable} and ModelScopeT2V~\cite{wang2023modelscope} have further improved accessibility and performance, enabling high-resolution video synthesis with enhanced temporal coherence.
In contrast, DiT-based methods, exemplified by models such as Sora~\cite{liu2024sora}, Open-Sora~\cite{opensora}, Vidu~\cite{bao2024vidu}, and CogVideoX~\cite{yang2024cogvideox}, leverage transformer architectures that operate on spatiotemporal patches of video latent codes. This strategy facilitates the generation of longer, higher-resolution videos with improved coherence and dynamism by effectively capturing long-range dependencies in both spatial and temporal dimensions. The emergence of diffusion transformers and autoregressive models highlights the increasing demand for large-scale, high-quality data, which this paper addresses within the human domain.

\begin{figure*}[htbp]
    \centering
    \includegraphics[width=0.93\textwidth]{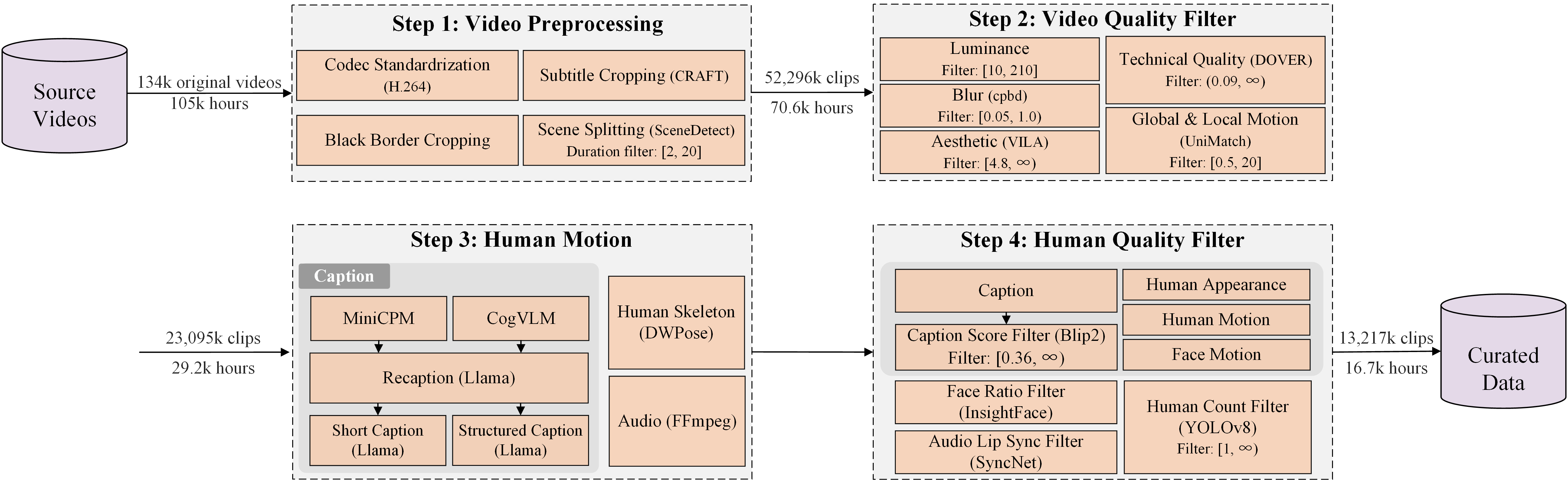} 
    % \vspace{-2.5mm}
    \caption{
The data processing pipeline.
The inputs are 105K hours of raw data from films, television shows, and documentaries, 
and the outputs are filtered high-quality videos that include textual captions—both short and long, as well as structured captions containing human information—and specific motion conditions related to individuals, such as skeleton sequences and speech audio. 
This pipeline consists of four key steps: video preprocessing, which involves basic decoding, cropping, and segmentation of the video; 
and video quality filtering, which assesses various quality metrics including luminance, blur, aesthetics, motion and technical quality. 
Then the human skeleton and speech audio are extracted from the video clips. 
An initial textual caption are generated with MiniCPM and CogVLM, voting by BLIP2 and reorgnized by the classic Llama model to obtain textual captions of different types.
Furthermore, the pipeline incorporates an advanced human quality filtering stage that aligns textual captions with the appearance, expressions, and pose movements of individuals, promoting fine-grained alignment between the textual information and the visual characteristics of the subjects.}
\label{fig:data_pipeline}
% \vspace{-mm}
\end{figure*}

\noindent\textbf{General Video Dataset.}
High-quality video-language datasets are crucial for advancing general video generation tasks. 
Notable large-scale datasets, such as HD-VILA-100M~\cite{xue2022advancing} and HowTo100M~\cite{miech2019howto100m}, utilize automatic speech recognition (ASR) for annotation; however, the captions generated often fail to adequately represent the video content, thereby limiting their utility for training purposes. 
WebVid-10M~\cite{bain2021frozen} stands out as a pioneering open-scenario text-to-video dataset, comprising 10.7 million text-video pairs with a cumulative duration of 52,000 hours. 
Panda-70M~\cite{chen2024panda} assembles 70 million high-resolution video samples characterized by strong semantic coherence. 
InternVid~\cite{wang2023internvid} leverages large language models to autonomously construct a video-text dataset, resulting in the generation of 234 million video clips accompanied by text descriptions. 
OpenVid-1M~\cite{nan2024openvid} provides expressive captions and curates 433,000 1080P videos, and its high-quality subset, OpenVidHD-0.4M, enhances high-resolution video generation.
Koala-36M~\cite{wang2024koala} includes 36 million video clips at a resolution of 720P, offering structured captions that average 200 words per segmented video clip, thereby improving the alignment between text and video content. 
Despite the advancements represented by these datasets, they exhibit a relative lack of diversity in terms of identity and motion within the human category, as well as deficiencies in high-quality textual captions, skeleton representations, and speech audio as motion conditions.

% \noindent\textbf{Human Video Dataset.}
% To support human image animation tasks, diverse and large-scale human-centric video datasets are essential. Real-world datasets collected from the Internet, such as the TikTok~\cite{jafarian2021learning}, contains over 300 dance video sequences shared on a social media, totaling more than 100,000 images. To expand dataset sizes cost-effectively, several synthetic datasets have been introduced. For example, SURREAL~\cite{varol2017learning} is a large-scale dataset featuring synthetically-generated but realistic images of people rendered from 3D sequences of human motion capture data. HSPACE~\cite{bazavan2021hspace} generates animations by fitting a human body model, offering diversity in characters, motions, and scenes. SynBody~\cite{yang2023synbody} is a large-scale synthetic dataset for human perception and modeling, sampling 10,000 animatable subjects. BEDLAM~\cite{black2023bedlam} renders people in realistic scenes and uses physics simulations to achieve realistic clothing effects. HumanVid~\cite{wang2024humanvid} introduces a synthetic dataset combined with a high-quality real-world video dataset, specifically designed for human image animation.
% While these datasets excel in specific tasks, they are not well-suited for pre-training or fine-tuning general video generative models in the human domain.
\noindent\textbf{Human Video Datasets.}
Diverse and large-scale human-centric video datasets are essential for advancing human image animation tasks. Real-world datasets collected from the Internet, such as TikTok~\cite{jafarian2021learning}, contain over 300 dance video sequences shared on social media platforms, totaling more than 100,000 images. UCF101~\cite{soomro2012ucf101} is a widely used action recognition dataset comprising realistic action videos collected from YouTube, organized into 101 action categories. NTU RGB+D~\cite{shahroudy2016ntu} is a large-scale dataset specifically designed for human action recognition, consisting of 56,880 samples spanning 60 action classes. Similarly, ActivityNet~\cite{caba2015activitynet} offers samples from 203 activity classes, totaling of 849 hours of video content. BEDLAM~\cite{black2023bedlam} generates realistic scenes of people using physics simulations to create accurate clothing effects, while HumanVid~\cite{wang2024humanvid} integrates synthetic and high-quality real-world video datasets tailored specifically for human image animation. Despite the utility of these datasets for specific tasks, they remain inadequate for pre-training or fine-tuning general video generative models in the human domain due to limitations in diversity, scale, and the specific motion conditions required for such models.

%% file: sections/dataset_arxiv.tex
% \begin{figure}[!t]
%     \centering
%     \includegraphics[width=1.0\linewidth]{figs/caption_demo5.png} 
%     \vspace{-6mm}
%     \caption{The illustration of output textual caption types and intermediate texts for caption selection.
%     %The output encompasses short, long, and structured captions specifically designed to address human characters. 
%     %The intermediate results include the original captions generated by CogVLM and MiniCPM, and the recaptioned texts employed in the selection process. 
%     The deep red text represents key information.}
%     \vspace{-7mm}
%     \label{fig:caption_demo} 
% \end{figure}

\begin{figure*}[!t]
    \centering
    \begin{subfigure}{0.33\linewidth}
        \centering
        \includegraphics[width=0.95\textwidth]{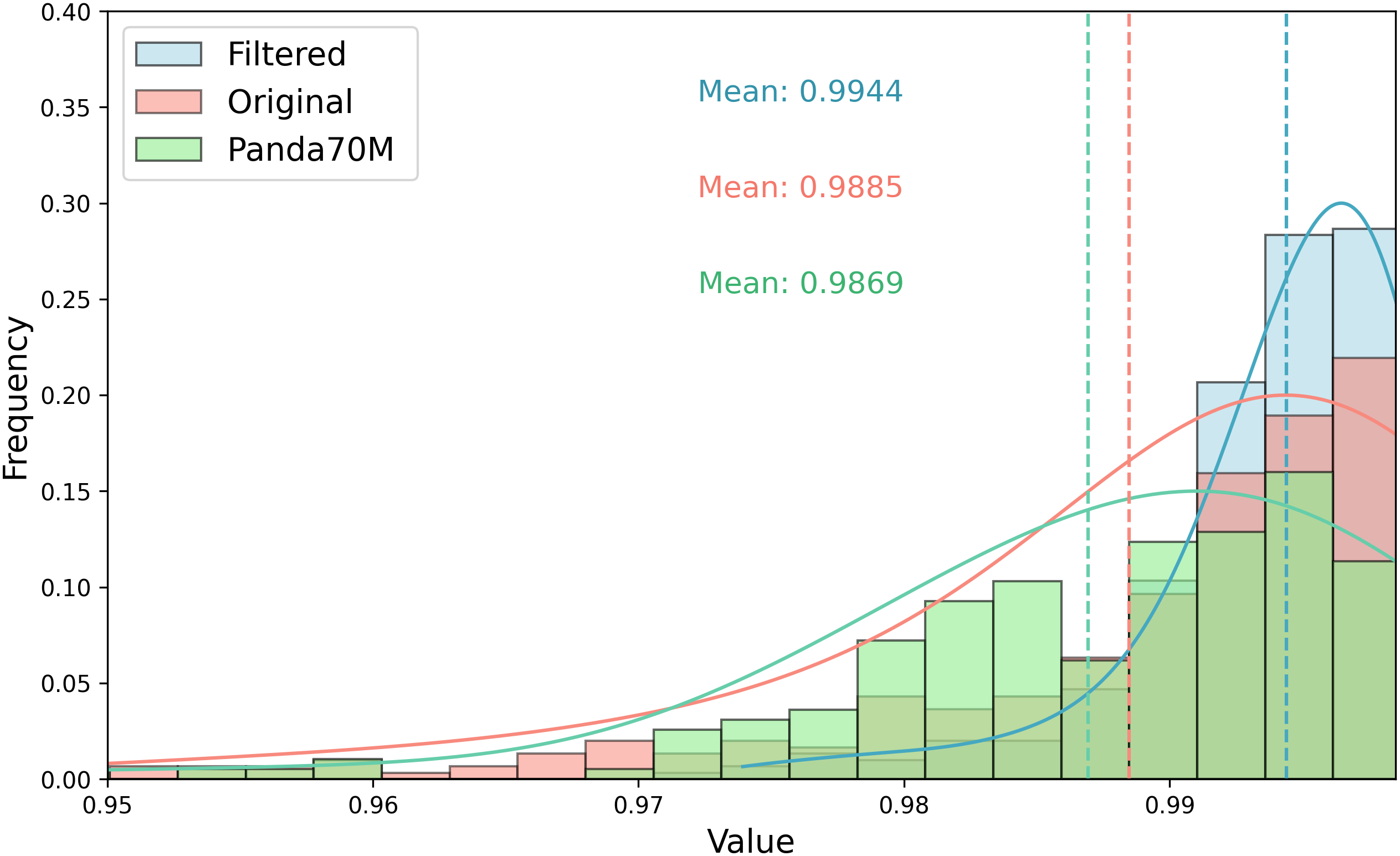} 
        \vspace{-1mm}
        \caption{Subject consistency ($\uparrow$)}  
        % \vspace{1mm}
        \label{fig:Vbench metirc}  
    \end{subfigure}%
    \hfill
    \begin{subfigure}{0.33\linewidth}
        \centering
        \includegraphics[width=0.95\textwidth]{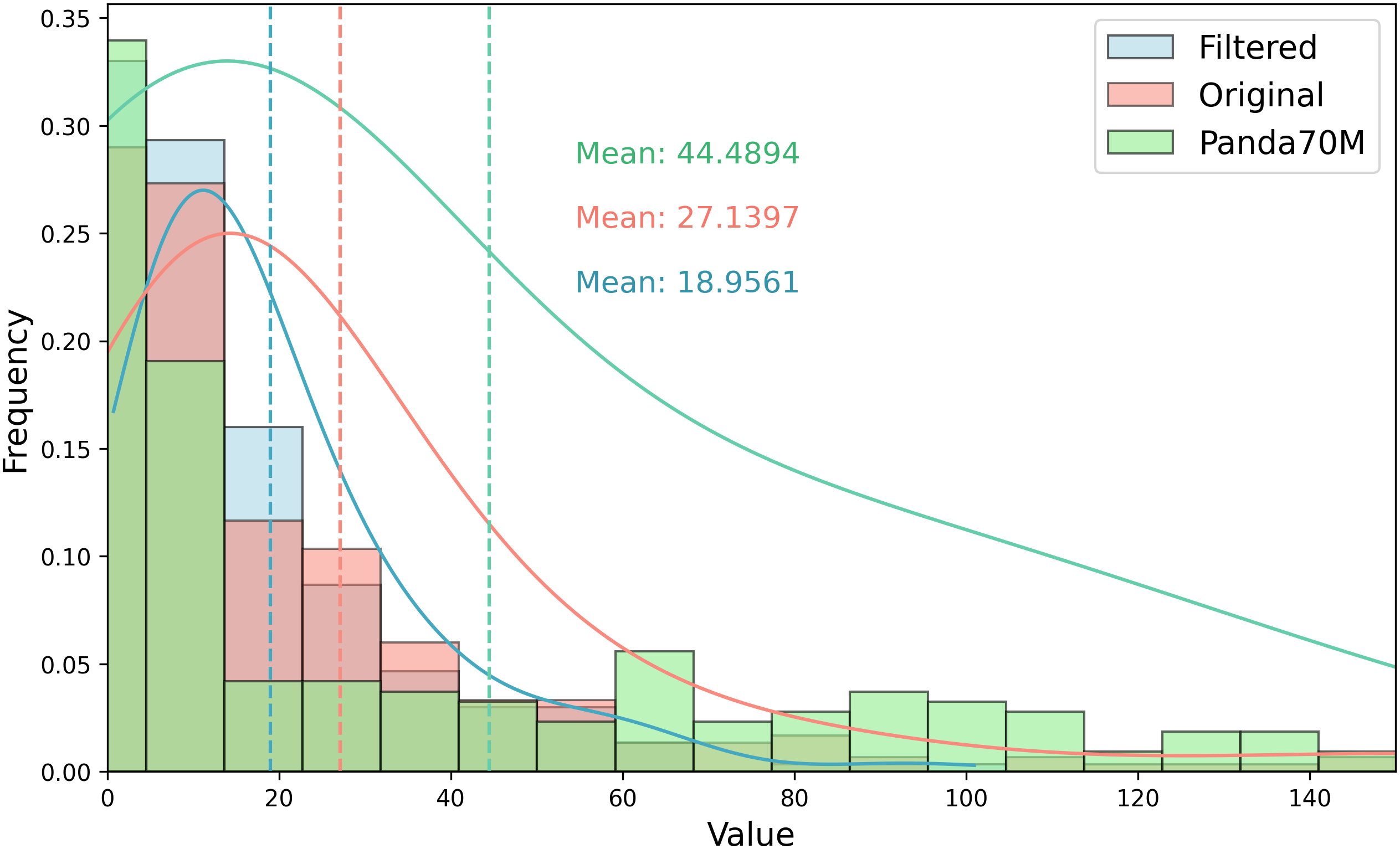}  
        \vspace{-1mm}
        \caption{Dynamic degree}  
        % \vspace{1mm}
        \label{fig:Vbench metirc}  
    \end{subfigure}%
    \hfill
    \begin{subfigure}{0.33\linewidth}
        \centering
        \includegraphics[width=0.95\textwidth]{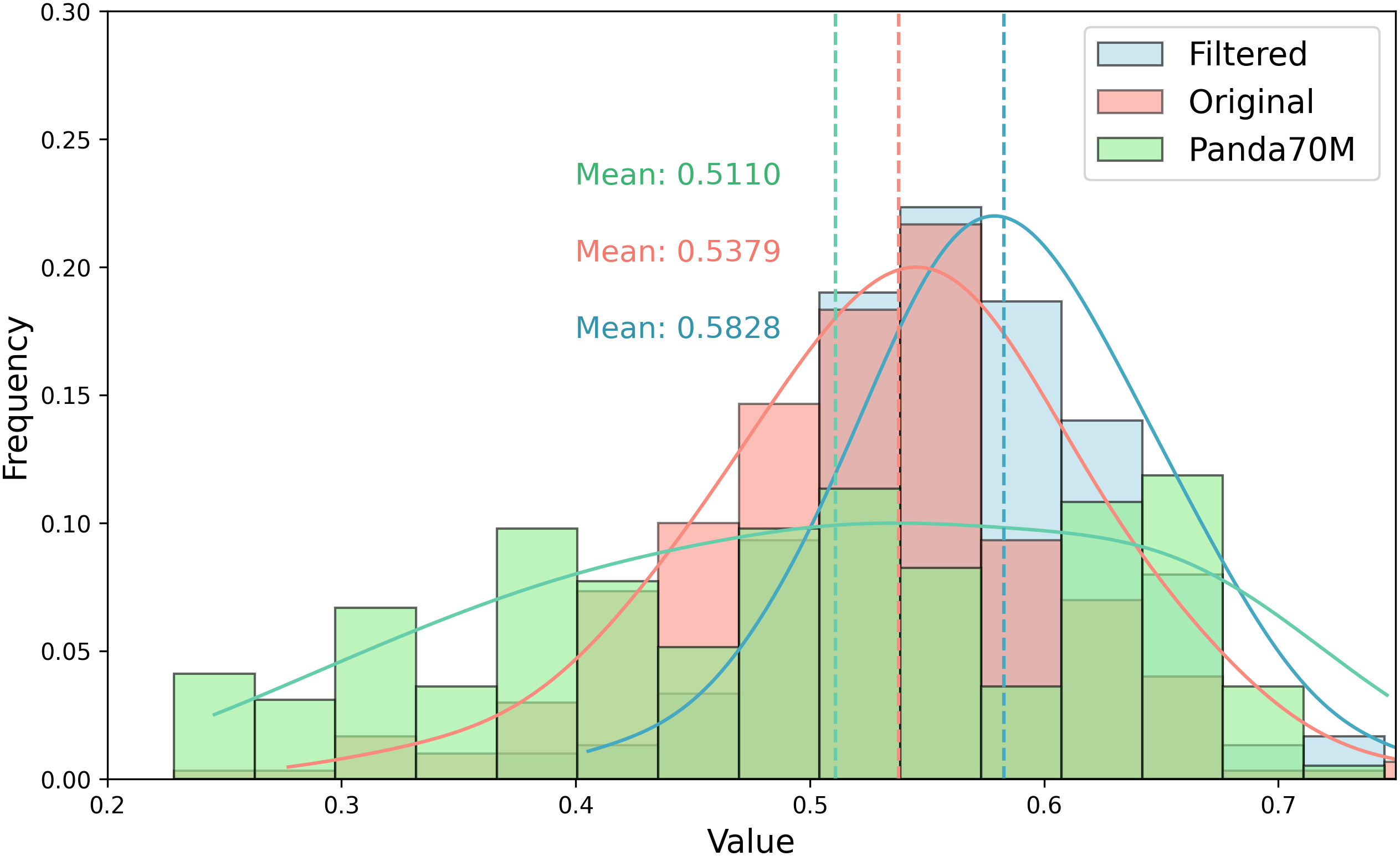}  
        \vspace{-1mm}
        \caption{Aesthetic quality ($\uparrow$)} 
        % \vspace{1mm}
        \label{fig:Vbench metirc}  
    \end{subfigure}  
    \begin{subfigure}{0.33\linewidth}
        \centering
        \includegraphics[width=0.95\textwidth]{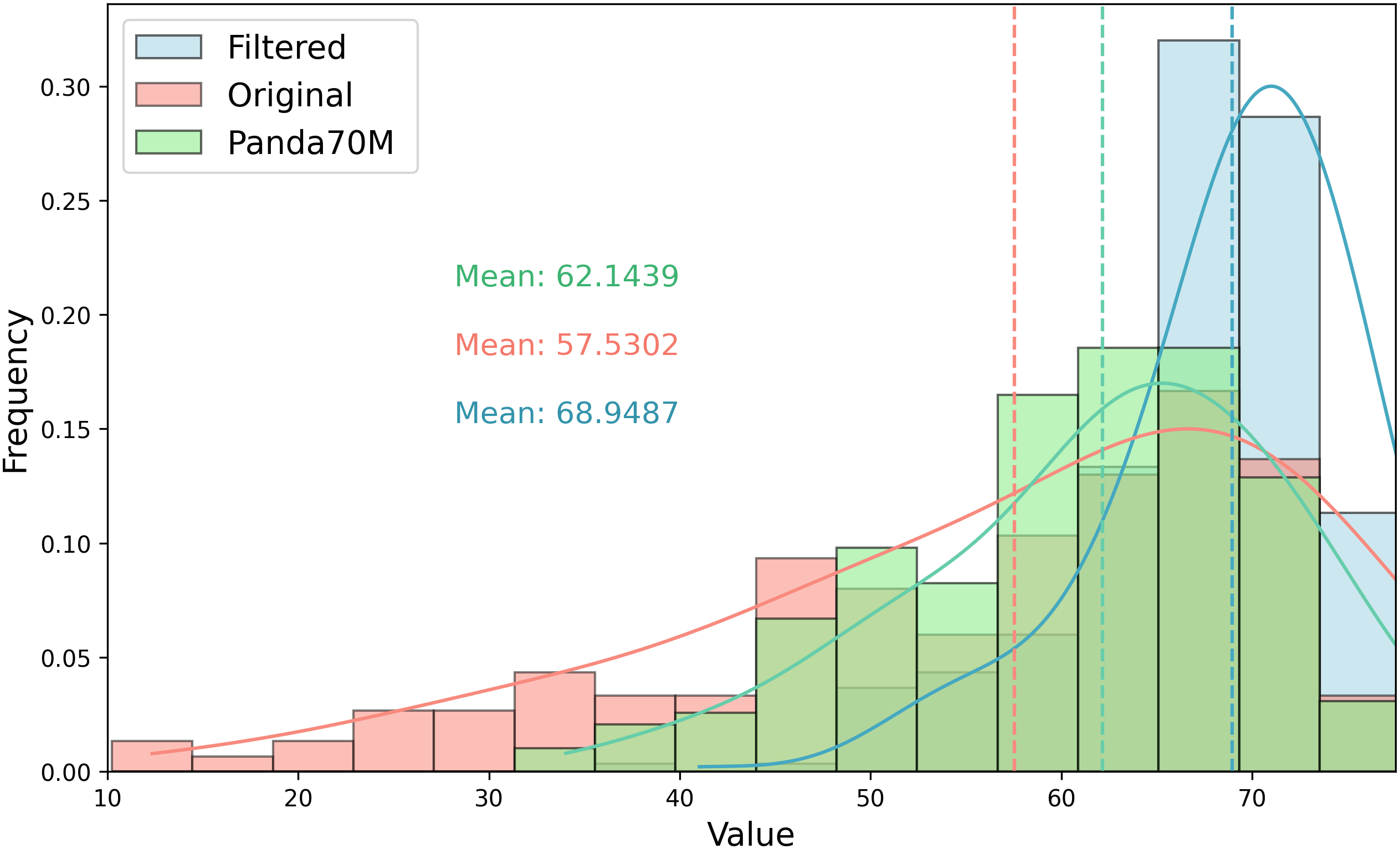}  
        \vspace{-1mm}
        \caption{Image quality ($\uparrow$)}
        % \vspace{1mm}
        \label{fig:Vbench metirc}  
    \end{subfigure}%
    \hfill
    \begin{subfigure}{0.33\linewidth}
        \centering
        \includegraphics[width=0.95\textwidth]{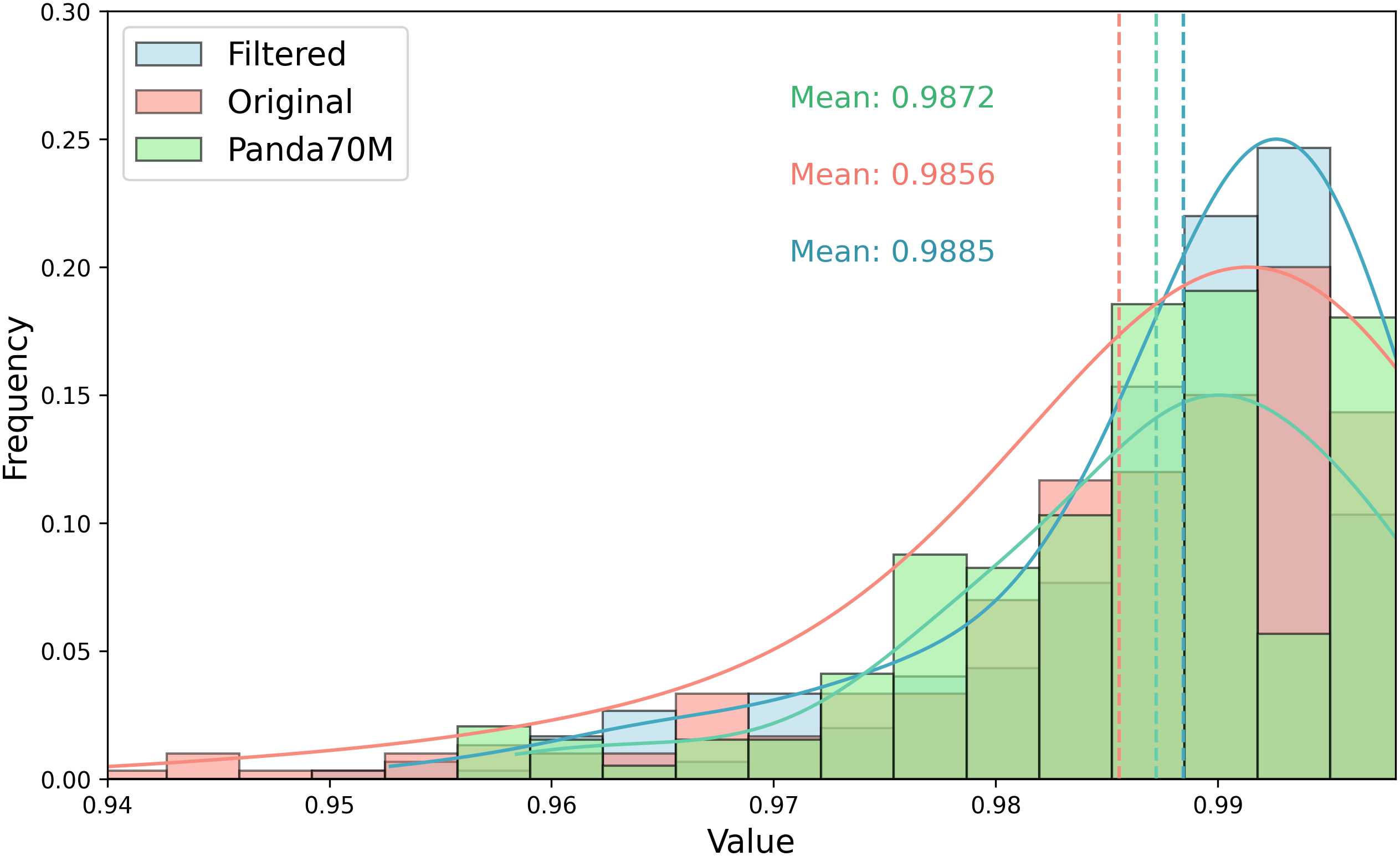}  
        \vspace{-1mm}
        \caption{Temporal flickering ($\uparrow$)}  
        % \vspace{1mm}
        \label{fig:Vbench metirc}  
    \end{subfigure}%
    \hfill
    \begin{subfigure}{0.33\linewidth}
        \centering
        \includegraphics[width=0.95\textwidth]{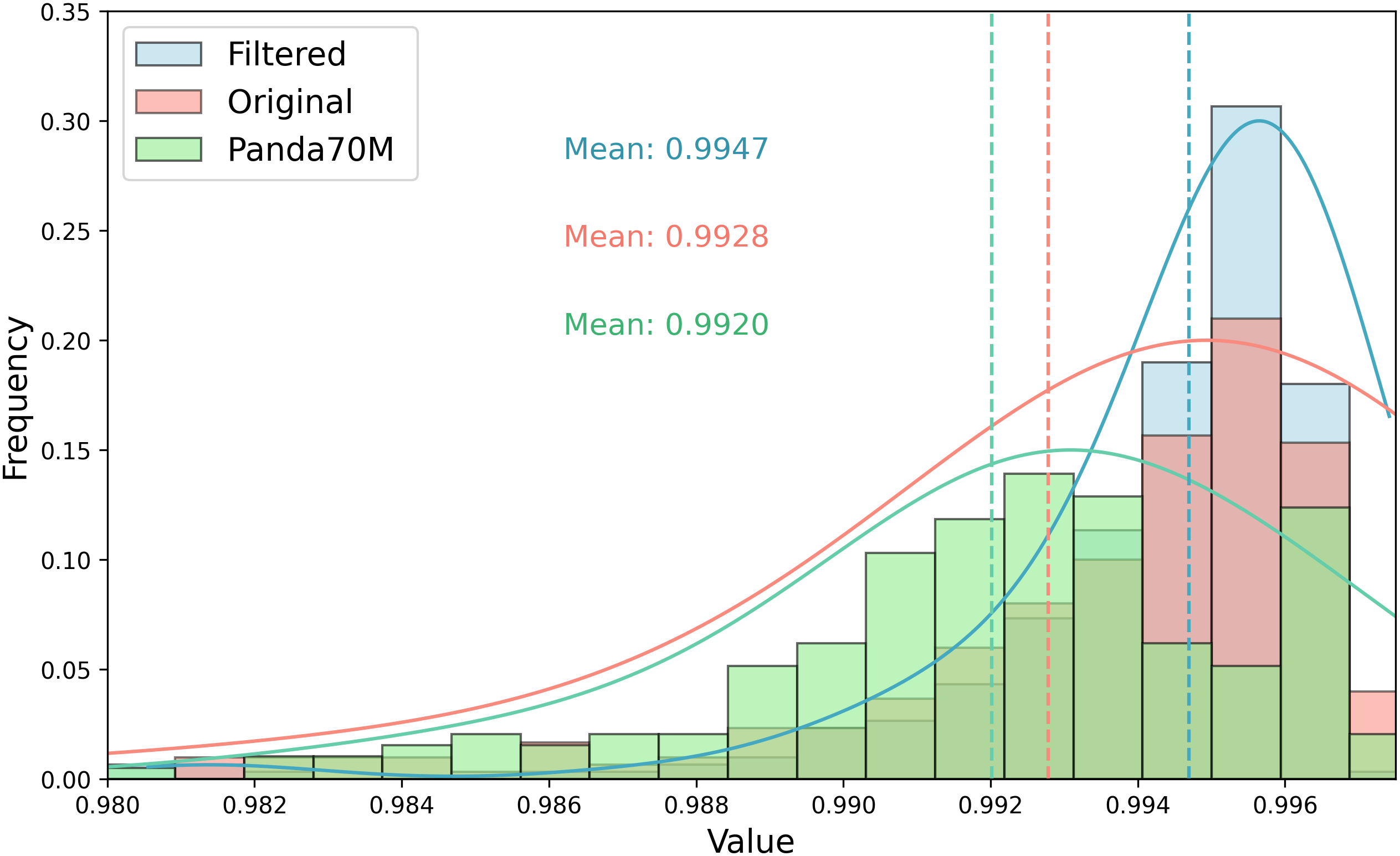}  
        \vspace{-1mm}
        \caption{Motion smoothness ($\uparrow$)}  
        % \vspace{1mm}
        \label{fig:Vbench metirc}  
    \end{subfigure}  
    % \vspace{-4mm}
    \caption{The comparison of video quality between Panda-70M and the proposed data before and after video quality filtering.
     We utilize the video quality evaluation metrics introduced in VBench to assess the video quality. 
     We can see that the general video quality of the proposed data has obviously improved after video quality filtering and superior to that of the Panda-70M.}
     \label{fig:quality}
     % \vspace{-5mm}
\end{figure*}

\section{Dataset}
% This section begins with a discussion of the source dataset in Section~\ref{sec:source_data}. 
% Section~\ref{sec:data_process} outlines the data processing steps, the generation of human motion, and a quality filter that considers the alignment between textual captions and human appearance, human motion, and face motion, as well as the synchronization between speech audio and human talking video. 

\subsection{Source Data}\label{sec:source_data}
As shown in Figure~\ref{fig:source_data}, the dataset comprises a total of 134,000 audiovisual works, with an aggregate runtime of 105,000 hours. This collection includes 5,192 films, 4,289 television series, and 192 documentaries, spanning a diverse range of genres across 15 categories, including Action, Adventure, Animation, Comedy, Crime, Documentary, Drama, Fantasy, Kids, Mystery, Politics, Romance, Science Fiction, Soap and War. The temporal scope of the dataset extends from the 1920s to the present day, featuring content in 58 different languages.
Table~\ref{tab:datasets} presents a comparative analysis of our dataset against previous general and human video datasets. The proposed human-centric dataset achieves a scale comparable to several well-known general video datasets while incorporating motion conditions typically found in human-specific datasets.
In terms of technical specifications, 84.5\% of the videos possess a resolution exceeding 720P, while 52.7\% exceed 1080P. 
% Furthermore, 74.0\% of the data exhibits a frame rate greater than 24 FPS, with an average bitrate surpassing 3,500 kbps. 

We selected films, television series, and documentaries as our primary data sources due to their emphasis on character-driven narratives, which typically exhibit superior luminance, aesthetic quality, and motion characteristics. 
Figure~\ref{fig:quality} demonstrates that the video quality after preprocessing is superior when compared to the Panda-70M dataset, which is primarily sourced from the internet.

\subsection{Data Processing}\label{sec:data_process}
\begin{figure}[!t]
    \centering
    \includegraphics[width=1.0\linewidth]{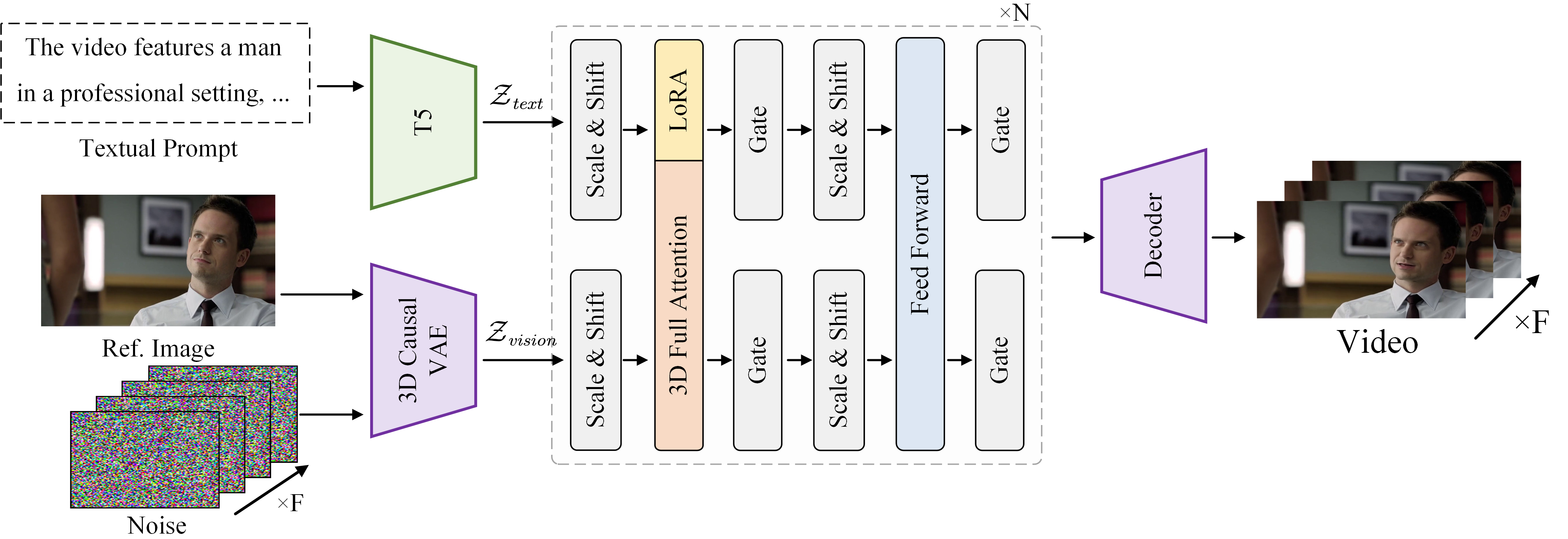} 
    % \vspace{-6mm}
    \caption{Overview of the proposed extended DiT-based video generation models.}
    \label{fig:network_pipeline}
    \vspace{-2mm}
\end{figure}

\subsubsection{Video Preprocessing}
The initial phase of video preprocessing encompasses several essential operations aimed at enhancing data quality. 
This phase includes codec standardization using H.264 and subtitle cropping via the CRAFT method~\cite{baek2019character}, which effectively removes spatial regions associated with subtitles.
Additionally, black border cropping is performed by analyzing pixel values across each frame and row to eliminate unwanted borders. Scene splitting is executed using SceneDetect~\cite{PySceneDetect}, which assesses color and brightness variations to identify significant transitions and determine potential split points. Finally, a duration filter is applied to trim video segments to a specified length of 2 to 20 seconds, ensuring the retention of relevant content for further analysis.

\subsubsection{Video Quality Filter}

\begin{figure*}[!t]
    \centering
    \includegraphics[width=1.0\linewidth]{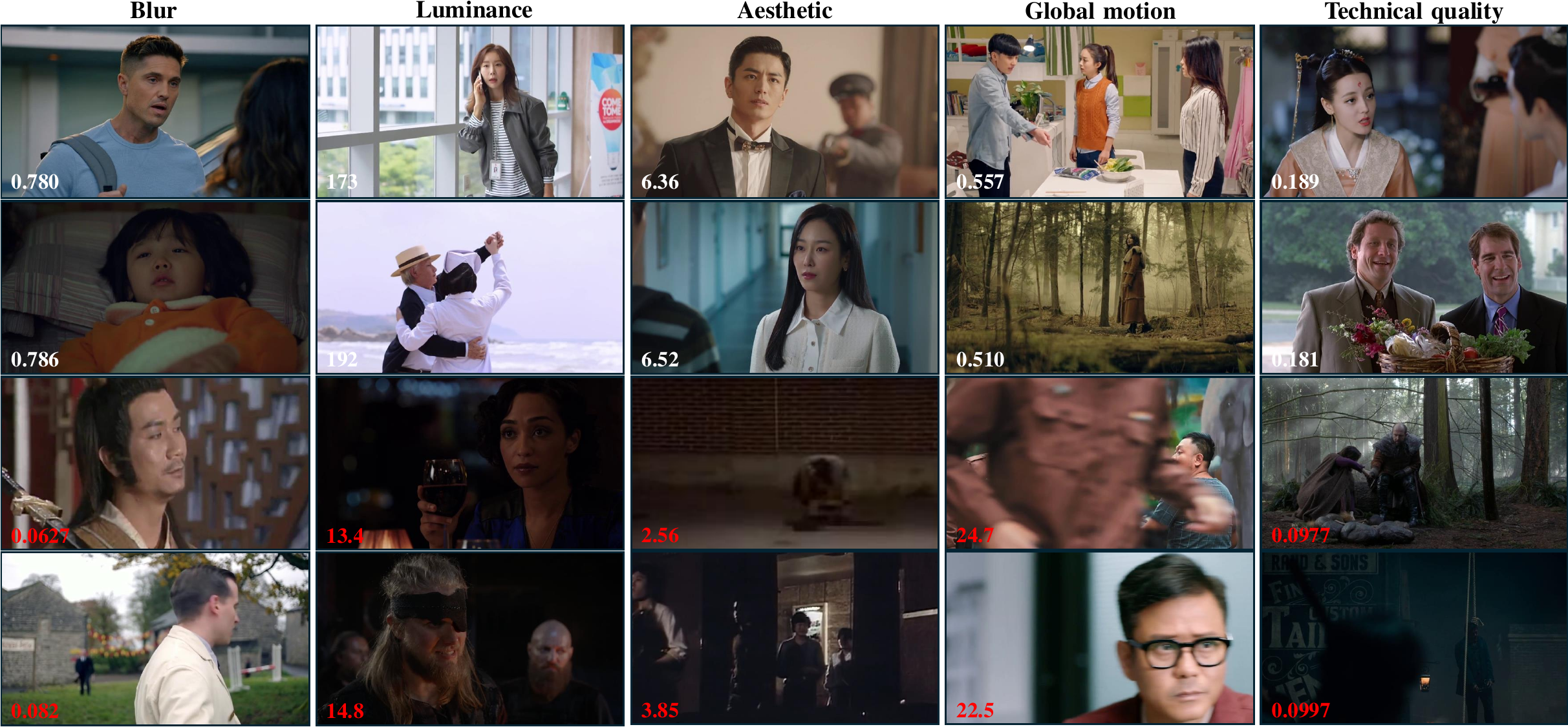}
    \caption{Videos we keep and deleted based on different quality filters. The number in the bottom-left corner of every image indicates the video's score for the corresponding quality filter. White numbers mean the video's score exceeds the threshold for this quality filter and the video is kept, while red numbers indicate the score is below the threshold and the video is deleted.}
    \label{fig:quality_filter}
    \vspace{-6mm}
\end{figure*}

In this study, we implement a multi-faceted video quality filtering process based on several key metrics: luminance, blurriness, aesthetic appeal, global and local motion, and overall technical quality. Each metric is evaluated against specific thresholds to ensure that only high-quality video content is retained for further processing.
Please see Figure~\ref{fig:quality} for the comparison of video quality before and after this filtering process.

\noindent\textbf{(1) Luminance.}  
Luminance is calculated using the formula: \(0.2126 \times R + 0.7152 \times G + 0.0722 \times B\), where \(R\), \(G\), and \(B\) denote the pixel values of the red, green, and blue channels, respectively. Videos are retained if their luminance values fall within the range of [10, 210].

\noindent\textbf{(2) Blur.}  
The level of blur in the video is assessed via the Cumulative Probability of Blur Detection algorithm \cite{narvekar2011no}, which analyzes edge feature distributions. Frames are extracted from the video, converted to grayscale, and the average, maximum, and minimum blur values are computed. Videos are filtered based on blur values within the range of [0.05, 1.0).

\noindent\textbf{(3) Aesthetic Quality.}  
Aesthetic appeal is measured using the CLIP-based Aesthetics Predictor \cite{aesthetic_predictor}. This method excludes videos with aesthetic scores below 4.8, thereby ensuring a satisfactory visual experience.

\noindent\textbf{(4) Global and Local Motion.}  
Motion quality is evaluated using optical flow analysis, specifically through the UniMatch algorithm \cite{unimatch}, which provides scores for both local and global motion. The filtering criteria for these scores are set within the range of [0.5, 20].

\noindent\textbf{(5) Technical Quality.}  
Lastly, the overall technical quality of the video is assessed using the DOVER model \cite{wu2023exploring}, filtering out videos with scores below 0.09.

Following these quality filters, we demonstrate the videos we keep and deleted in the filtering process in Figure~\ref{fig:quality_filter}.

\begin{figure*}[!htbp]
    \centering
    \includegraphics[width=1.0\linewidth]{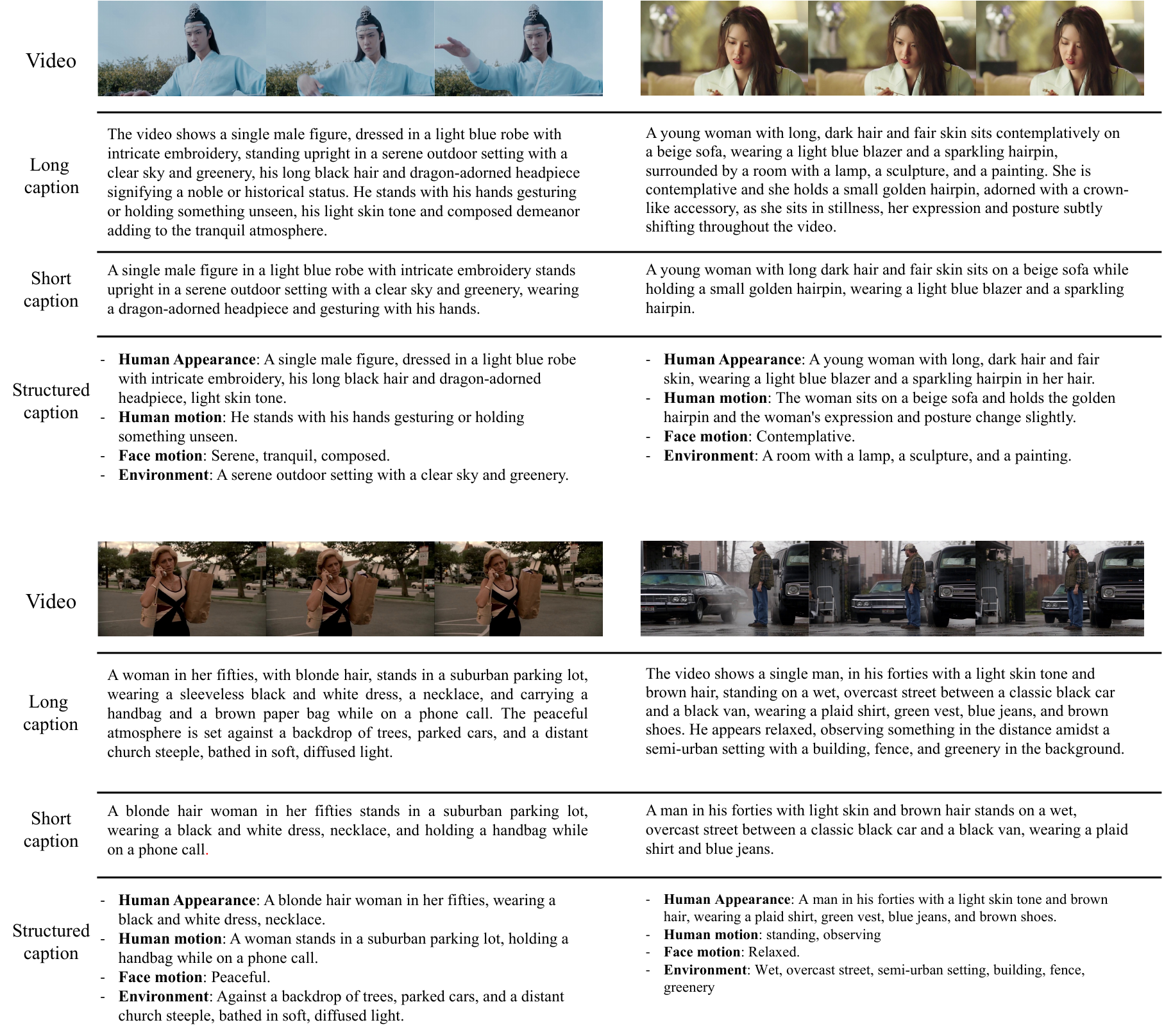} 
    \caption{The illustration of textual captions and corresponding types (long, short, and structured captions). }
    \label{fig:supp_caption}
    % \vspace{-6mm}
\end{figure*}

% \begin{figure*}[!htbp]
%     \centering
%     \includegraphics[width=1.0\linewidth]{figs/supp/supp_skeleton.pdf} 
%     \caption{The illustration of human skeleton sequences with respect to given videos.}
%     \label{fig:supp_skeleton} 
%     \vspace{+6mm}
% \end{figure*}

% \begin{figure*}[!htbp]
%     \centering
%     \includegraphics[width=1.0\linewidth]{figs/supp/supp_speech.pdf} 
%     \caption{The illustration of speech audio with respect to given videos. Left: video screenshot; right: speech script.}
%     \label{fig:supp_speech} 
%     \vspace{-6mm}
% \end{figure*}

\subsubsection{Human Motion Generation}
\label{sec:human_motion}
\noindent\textbf{Textual Caption.}
% In video generation models, textual captions play a critical role by providing contextual information that guides the generation process, enhances multimodal learning, and improves model generalization. 
% They encourage alignment between generated visual content and textual descriptions.
As illustrated in Figure~\ref{fig:supp_caption}, we employ two publicly available multimodal models, namely MiniCPM~\cite{yao2024minicpm} and CogVLM~\cite{hong2024cogvlm2}, to generate textual captions for the provided videos. 
MiniCPM has been extensively utilized in prior research~\cite{chen2024multi}, while CogVLM is a more recent introduction to the field. 
To enhance the quality of the generated captions, we implement a voting strategy. 
However, the limited instruction-following capabilities of these models often lead to captions with chaotic structures that lack a unified format, thereby complicating the training of video generation models. 
To address this issue, we utilize LlaMA 3.1~\cite{dubey2024LlaMA} to reformat the initial captions. 
Subsequently, we employ BLIP2~\cite{li2023blip} to select captions with higher similarity scores from the two generated sets.

The final captions are provided in short, long and structured formats to ensure a comprehensive representation of the video content. 
Long captions are derived from the aforementioned process. 
To obtain structured captions, we utilize LlaMA to extract components related to human appearance, human motion, face motion and environment description. 
Finally, we obtain the short captions by merging the structured captions by LlaMA and restricting caption length to approximately 20 to 30 words.

\noindent\textbf{Skeleton Sequence.}  
The adoption of skeleton poses has become prevalent in diffusion-based human video generation models~\cite{zhu2024champ,chang2023magicpose,hu2024animate}. 
In our dataset, we utilize DWpose to extract skeleton data, which serves as the foundational annotation for controlling human poses in the generated videos. 
The human skeleton sequences obtained through this method, along with the corresponding human videos, aim to enhance the existing human animation datasets by contributing to the diversity of characters, scenes, and movements.

\noindent\textbf{Speech Audio.}  
Recent studies have investigated audio-driven head animation ~\cite{xu2024hallo,xu2024vasa,he2023gaia}, gesture generation ~\cite{yang2023qpgesture,zhu2023taming}, and holistic human motion ~\cite{corona2024vlogger,wang2024dance}, underscoring the significance of audio as a guiding modality. 
In our dataset, we include the corresponding audio tracks for each video clip. 
Furthermore, we employ SyncNet~\cite{raina2022syncnet} to assess the consistency between the speech audio of the subject and their lip motion. 
We anticipate that the integration of speech audio in human videos will further advance research in audio-driven animation, extending its applicability to more realistic and generalized scenarios.

\subsubsection{Human Quality Filter}
\label{sec:caption_refine}

This phase focuses on assessing the presence and characteristics of human subjects within the video. Key criteria for filtering include the number of individuals detected, the dimensions of human and facial regions, and the velocity of human motion. Additionally, the alignment of text and video concerning human appearance, motion, and facial expressions is also considered.

\noindent\textbf{Human Appearance Text-Video Alignment.}
We assess the alignment between the generated video and the human appearance descriptions provided in the text prompt, which include characteristics such as physical features, style, clothing, and accessories. 
By extracting the appearance-related text from the generated long caption, we compute the BLIP2 score~\cite{li2023blip} to evaluate the correspondence between the appearance text and the generated video. 
This process ensures that the visual attributes of the human subjects are accurately represented in accordance with the textual descriptions.

\noindent\textbf{Human Motion Text-Video Alignment.}
We evaluate the alignment of the generated video with the human motion described in the text prompt, which encompasses aspects such as action type, speed, and trajectory. 
Utilizing the LlaMA model, we extract the action-related text from the original caption and compute the BLIP2 score~\cite{li2023blip} to assess the semantic alignment between the action text and the generated video. 
This evaluation promotes the accurate depiction of actions in the video as intended by the descriptions.

\noindent\textbf{Face Motion Text-Video Alignment.}
Similarly, we analyze the extent to which the generated video aligns with facial motion, including expressions, head posture, and emotional variations described in the text prompt. 
By extracting the emotion-related text using the LlaMA model, we calculate the BLIP2 score~\cite{li2023blip} to ensure that the emotional content is accurately conveyed. 
This alignment is essential for maintaining the intended emotional narrative of the video.

%% file: sections/network.tex
\section{Network}

\noindent\textbf{Baseline Diffusion Transformer Network.}
The proposed baseline transformer diffusion network is based on the CogVideoX architecture~\cite{yang2024cogvideox} and utilizes a causal 3D VAE~\cite{kingma2022autoencoding} for video compression, achieving temporal and spatial factors of 4 and 8, respectively. Latent variables are formatted as sequential inputs, while textual inputs are encoded into embeddings using the T5 model~\cite{raffel2020exploring}. These inputs are processed together in a stacked Expert Transformer network, incorporating Adaptive Layer Normalization for alignment and 3D RoPE~\cite{su2024roformer} to improve the model's capacity for temporal dynamics and long-range dependencies within video frames.

\noindent\textbf{Extension to Baseline.} 
In light of our dataset, our objective is to enhance the video generation model's capability to produce human subjects while maintaining identity appearance and generating facial expressions and human motions that accurately and naturally correspond to textual captions. However, when the pretrained baseline model is further trained on new data, it risks overfitting to the new task, which can lead to catastrophic forgetting. This phenomenon results in the model losing the broad knowledge acquired during the pretraining phase.

To address this challenge, we adopt the Low-Rank Adaptation (LoRA) technique. 
Specifically, LoRA~\cite{hu2021lora} targets the residual components of the model, denoted as \(\Delta W\), which is incorporated into the original weight matrix as follows:
$W' = W + \Delta W$.
Here, \(\Delta W\) is expressed as the product of two low-rank matrices:
$\Delta W = AB^T, \quad A \in \mathbb{R}^{n \times d}, \, B \in \mathbb{R}^{m \times d}, \, d < n, \, d < m$.
By focusing on the smaller matrices \(A\) and \(B\) rather than the full weight matrix \(W\), LoRA significantly reduces the computational and memory overhead during training. 

As shown in Figure~\ref{fig:network_pipeline}, this technique leverages the proposed data to train the new parameters, thereby circumventing the need to retrain all model parameters and enhancing the model's ability to generate high-quality human representations aligned with textual descriptions.

%Through the implementation of LoRA, illustrated in Figure~\ref{fig:network_pipeline}, we aim to retain the valuable knowledge acquired during pretraining while effectively adapting to new data, thereby enhancing the model's ability to generate high-quality human representations aligned with textual descriptions.

%% file: sections/experiments_arxiv.tex
\begin{figure*}[th!]  % 使用figure环境
    \centering
    \includegraphics[width=1.0\textwidth]{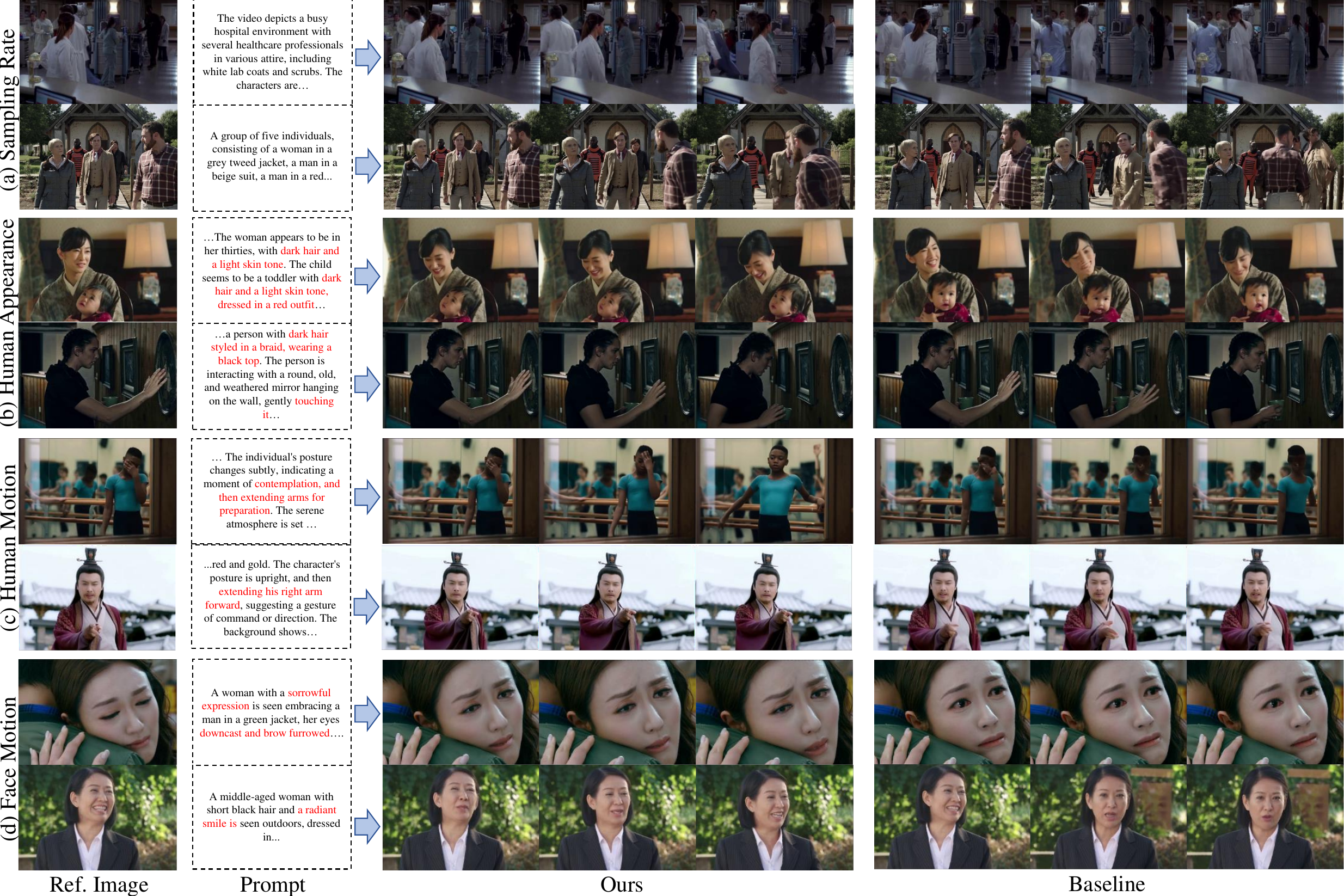}  
    % \vspace{-7mm}
    \caption{Visual comparison among four training strategies. (a) High video sampling rates (first and second rows) prevent appearance degradation during quick movements, while low FPS leads to significant visual inconsistencies.(b) The human appearance filter (third and fourth rows) eliminates facial blurring and deformities in faces and hands.(c) Our method aligns better with action prompts like "extending arms" (fifth and sixth rows), unlike the baseline which fails to follow the prompts.(d) Our method accurately reflects facial expressions prompted, such as "sorrowful expression" and "smile" (seventh and eighth rows), whereas the baseline does not.}
    \label{fig:ablation_compare} 
    % \vspace{-5mm}
\end{figure*}

\section{Experiments}
\subsection{Implementation}

\noindent\textbf{Training.}
During the training phase, we utilized H100 GPU server cluster and incorporated low-rank matrices with a rank of 1024 into the full 3D attention module of the baseline network. 
After incorporating LoRA, the total number of parameters in the model was 6.60 billion, with 1.02 billion trainable parameters. We froze the gradients of all weights in the original base network and trained for 20,000 steps with a learning rate of 5e-4, implementing both learning rate warm-up and decay mechanisms. 

\noindent\textbf{Inference.}
During the inference phase, we employed 50 diffusion steps and dynamic Classifier-Free Guidance (CFG)~\cite{ho2021classifier} to enhance the generation process. The generated videos have a resolution of 720$\times$480 and consist of 49 frames per sequence. This setting is consistently applied across all experiments to ensure a fair comparison.

\noindent\textbf{Data.}
To validate the effectiveness of our data and training strategy, we selected 6.55K hours from the filtered video dataset using a human quality filter and adopted a sampling rate of 24 FPS, an increase from the previous baseline of 8 FPS. 
%This training dataset ensures high video quality, specifically a higher DOVER value, and captures a broader range of human expressions and movements across diverse scenes.
Prior to large-scale training, we conducted small-scale experiments to evaluate various training data strategies. 
For these experiments, we randomly selected 1.05K hours from the 6.55K hour dataset and performed four comparative experiments on different training strategies: video sampling rate, human appearance text-video alignment, human motion text-video alignment, and face motion text-video alignment.
For the test set, we randomly selected 240 samples from the dataset that were not used during training. 
These samples encompass a variety of human proportions and scenes. Additionally, we selected 100 samples where humans occupy a significant spatial portion, aiming to better assess human evaluation metrics.

\begin{table}[t!]
\centering
\resizebox{\linewidth}{!}{
\begin{tabular}{@{}cccc@{}}
\toprule
Category& Metrics& Description \\
\midrule
\multirow{2}{*}{Human Consistency} & Face Consistency             & Evaluates the consistency of facial feature across the video.          \\
 & Body Consistency& Evaluates the consistency of the body feature across the video.         \\
\midrule
 \multirow{2}{*}{Human Semantics}&Body Semantics&Measures semantic alignment of human body motion and text prompt.\\
&Face Semantics&Measures semantic alignment of human face motion and text prompt.\\
\midrule
\multirow{4}{*}{Vbench}&I2V Consistency &Quantifies adherence to a reference image across frames. \\
&Aesthetic Quality &Scores artistic and beauty of video.\\
&Imaging Quality &Assesses visual technical quality. \\
&Motion Smoothness &Evaluates naturalness of motion. \\
\bottomrule
\end{tabular}}
% \vspace{-4mm}
  \caption{Metrics used for evaluating the generated results.}
  \label{tab:metrics}
  % \vspace{-7mm}
\end{table}

\begin{table*}[th!] 
  % \centering
\resizebox{\textwidth}{!}{
\begin{tabular}{c|cccccccc}
\toprule
\multirow{2}{*}{Video sample rate} & \multicolumn{2}{c}{Human Consistency} & \multicolumn{2}{c}{Human Semantics} & \multicolumn{4}{c}{VBench Evaluation}                                     \\ \cmidrule{2-9} 
                                   & Face Consistency  & Body Consistency & Body Semantics     & Face Semantics      & I2V Consistency & Aesthetic Quality & Imaging Quality & Motion Smoothness \\
                                   \cmidrule{1-9} 
FPS = 8                            & 0.7209            & 0.9540           & 0.4217          & 0.3021          & 0.9763          & 0.5546            & 0.6534          & 0.9929            \\
FPS = 24                           & \textbf{0.7390}   & \textbf{0.9638}  & \textbf{0.4262} & \textbf{0.3055} & \textbf{0.9825} & \textbf{0.5572}   & \textbf{0.6587} & \textbf{0.9946}   \\ \hline
\end{tabular}
  }
  % \vspace{-3 mm}
  \caption{Quantitative comparison of the extended diffusion transformer network further pretrained on the proposed 1.05K hours dataset with different video sampling strategy. We can see that using higher video sampling rate as train data markedly improves human appearance consistency, meanwhile enhancing the video quality.}
  \label{tab:fps_compare}
  % \vspace{-3 mm}
\end{table*}

\begin{table*}[th!]
\resizebox{\textwidth}{!}{
\begin{tabular}{c|cccccccc}
\toprule
\multirow{2}{*}{Filter Method} & \multicolumn{2}{c}{Human Consistency} & \multicolumn{2}{c}{Human Semantics} & \multicolumn{4}{c}{VBench Evaluation}                                     \\ \cmidrule{2-9} 
                               & Face Consistency  & Body Consistency & Body Semantics      & Face Semantics     & I2V Consistency & Aesthetic Quality & Imaging Quality & Motion Smoothness \\ 
   \midrule
No Filter                      & 0.7976            & 0.9679           & 0.4437           & 0.3148         & 0.9821          & 0.5656            & 0.6912          & 0.9941            \\
Human Appearance Filter        & 0.8674            & 0.9810           & 0.4556           & 0.3197         & 0.9898          & 0.5843            & 0.7077          & 0.9951            \\
Human Motion Filter            & 0.8275            & 0.9796           & 0.4866           & 0.3144         & 0.9888          & 0.5575            & 0.6971          & 0.9946            \\
Face Motion Filter             & 0.8407            & 0.9794           & 0.4454           & 0.3220         & 0.9891          & 0.5685            & 0.7007          & 0.9950            \\ \bottomrule
\end{tabular}
}
% \vspace{-3 mm}
\caption{Quantitative comparison of the extended diffusion transformer network further pretrained on the proposed 1.05K hours dataset with different human quality filter, including the filter of human appearance text-video alignment, human motion text-video alignment and face motion text-video aligment.  }
\vspace{+3 mm}
  \label{tab:quan_compare_furthertrain}
\end{table*}

\begin{table*}[th!] 
  \centering
  \resizebox{\textwidth}{!}{
  \begin{tabular}{c|cccccccc}
    \toprule
    & \multicolumn{2}{c}{Human Consistency} & \multicolumn{2}{c}{Human Semantics} & \multicolumn{4}{c}{VBench Evaluation} \\
    \cmidrule(r){2-3} \cmidrule(r){4-5} \cmidrule(r){6-9}
    & Face Consistency & Body Consistency & Body Semantics & Face Semantics & I2V Consistency & Aesthetic Quality & Imaging Quality & Motion Smoothness \\
    \midrule
    Baseline (CogVideoX) &0.7108 & 0.9405 & 0.4166&0.3018 & 0.9759  & 0.5562 & 0.6530 &0.9905\\
    Ours & \textbf{0.7418} & \textbf{0.9678} & \textbf{0.4333} & \textbf{0.3136} & \textbf{0.9885} & \textbf{0.5594} & \textbf{0.6662} & \textbf{0.9948}\\
    \bottomrule
  \end{tabular}}
  % \vspace{-3 mm}
  \caption{Quantitative comparison between the baseline and the proposed extended diffusion transformer network further pretrained on the proposed 6.05K hours dataset. }
  % \vspace{-6 mm}
  \label{tab:action_compare}
\end{table*}

\noindent\textbf{Evaluation Metrics.}
%There are several evaluation metrics for generated videos, such as Vbench~\cite{huang2024vbench}, which assesses video quality based on factors like subject appearance consistency, motion semantics, and low-level visual quality. 
In this experiment, to evaluate the general quality of the generated video results, we employed the following metrics from Vbench: Image-to-Video Consistency, Motion Smoothness, Aesthetic Quality, and Imaging Quality. 
%These metrics allow us to assess the visual continuity, smoothness of motion while adhering to physical laws, as well as the aesthetic appeal and technical distortions of the generated videos.
As shown in Table~\ref{tab:metrics}, we aim to focus on the generative performance of the human in the video. Inspired by Subject Consistency in Vbench, we introduce two metrics: \textbf{Face Consistency} and \textbf{Body Consistency}. Face Consistency evaluates whether facial features remain consistent across the video. We measure this by calculating feature similarity across frames using the InsightFace model~\cite{Deng2020CVPR}, which is specialized in face tasks. Body Consistency evaluates the consistency of the body in a similar manner, we employ the YOLOv8 detection model to extract human body, followed by DINO~\cite{caron2021emerging} for feature extraction and similarity comparison.
We also introduce two metrics to measure the semantic alignment of body motion and face motion: \textbf{Body Semantics} and \textbf{Face Semantics}. Specifically, we use the similarity score between the generated video and words of body motion and face motion as an evaluation metric for human-video alignment, in the following experiments, the BLIP2 score is employed.

% \begin{figure*}[th!]  % 使用figure环境
%     \centering
%     \includegraphics[width=1.0\textwidth]{figs/vis_res/generated_results.png}  
%     \vspace{-6mm}
%     \caption{Different visual generation results of ours diffusion transformer extended model. It can be seen from the first row and the second row, we can produce human more stable portrait videos, and anime characters with high character consistency. In addition to that, ours model also can handle complex motion, like the third row, person is riding a horse without any distortion.}  
    
%     \label{fig:generated_results} 
%     \vspace{-5mm}
% \end{figure*}

\subsection{Training Data Strategy}
\label{sec：training_strategy}
In this section, we build upon a prestigious baseline diffusion transformer network, enhancing it with the LoRA technique, and subsequently pretraining the network on the proposed dataset.
Based on the results of the further pretraining, we derive four key insights.

\noindent\textbf{Video sampling rate.}
The video sampling rate is critical for maintaining identity consistency in generated human videos and aligning facial and body movements with textual captions. 
A higher sampling rate facilitates the capture of more frames, enhancing temporal coherence in features such as facial expressions and body postures. 
Moreover, effective alignment of movements with textual descriptions hinges on the model's ability to accurately capture motion dynamics over time. 
An increased sampling rate provides detailed information about these dynamics, enabling realistic movements that correspond to captions.

Table~\ref{tab:fps_compare} illustrates the training strategy employed to enhance the video sampling rate. Results indicate that a frame rate of 24 FPS yields significant improvements in appearance consistency and in the alignment between human motion and text prompts, alongside enhancements in general video metrics. 
As depicted in Figure~\ref{fig:ablation_compare} (a), higher FPS markedly improves character appearance, particularly in complex scenes with numerous individuals, whereas lower FPS leads to discernible artifacts in human representations.

\noindent\textbf{Human Appearance Text-Video Alignment.} As shown in Table~\ref{tab:quan_compare_furthertrain}, the introduced human appearance text-video alignment method significantly improves both face consistency and body consistency, and slightly improves body semantics and face semantics, alongside improvements in general video metrics. As depicted in Figure~\ref{fig:ablation_compare}~(b), our method shows remarkable consistency in character appearance across frames, whereas the baseline version shows more visual degradation in appearance, such as the face and hands.

\noindent\textbf{Human Motion Text-Video Alignment.} As shown in Table~\ref{tab:quan_compare_furthertrain}, the introduced human motion text-video alignment method significantly improves body consistency and body semantics, alongside improvements in general video metrics, which indicates better adherence to the actions specified in the prompts. As depicted in Figure~\ref{fig:ablation_compare}~(c), our method shows higher adherence to the actions indicated in the prompts, while the baseline method failed to follow the prompts.

\noindent\textbf{Face Motion Text-Video Alignment.} As shown in Table~\ref{tab:quan_compare_furthertrain}, the introduced face motion text-video alignment method shows significant improvement in both face consistency and face semantics, alongside improvements in general video metrics. As depicted in Figure~\ref{fig:ablation_compare}~(d), our method shows higher adherence to the expressions indicated in the prompts, while the baseline method failed to follow the prompts.

% Our analysis indicates that the accuracy of textual captions regarding character appearance, facial expressions, and body motion significantly impacts the final outcomes. 
% The alignment between textual descriptions and video content is essential for achieving high-quality results.
% In light of these insights, we validate the effectiveness of the proposed human quality filter and further design a data training strategy for the model.

% Finally, based on our large and diverse dataset, combined with various training strategies, we find that, as shown in Table~\ref{tab:quan_compare_furthertrain}, our results significantly outperform those of baseline network in generating human videos. 

% As seen in the Figure~\ref{fig:ablation_compare} (b)(c)(d), the appearance filter prevents artifacts on the hands and face. Additionally, with the motion and expression filters, the human's actions and expressions better follow the prompt instructions, such as raising a hand, furrowing brows, or smiling.

\subsection{Training on OpenHumanVid} 
As shown in Table~\ref{tab:action_compare}, utilizing the proposed large-scale and high-quality human video dataset, and incorporate the training data strategies described above, including higher video sampling rate, human appearance filter, human motion filter and face motion filter, the extended model obviously improves compared with the baseline model after further pretraining, evaluated no matter in general video evaluation metrics and human-related evaluation metrics.

\subsection{Limitations and Future Works}
Despite the advancements presented by OpenHumanVid, several limitations warrant consideration. 
Firstly, the reliance on existing multimodal models for caption generation may restrict the diversity and richness of the descriptions, potentially impacting the comprehensiveness of video-text alignment. 
Additionally, while the dataset incorporates various human motion conditions, the inherent variability of human actions and expressions across different cultures and contexts may not be fully captured. 
Future work could focus on enhancing the caption generation process through the development of more sophisticated models with improved instruction-following capabilities. 
Furthermore, expanding the dataset to include a wider array of human identities and motion scenarios, along with more granular textual prompts, could facilitate the generation of videos that better reflect the complexities of human behavior and interaction in diverse contexts.

\subsection{Safety Considerations}
The development of OpenHumanVid introduces several social risks, particularly concerning privacy, representation, and the potential for misuse. The dataset's inclusion of diverse human identities and movements necessitates rigorous ethical considerations to prevent the reinforcement of stereotypes or biases in generated videos. Additionally, there is a risk of inappropriate usage in contexts such as deepfakes or unauthorized portrayals of individuals. To mitigate these issues, we implement strict data governance protocols, ensuring that all content adheres to ethical standards and is used responsibly. Furthermore, we advocate for transparency in the dataset's application, emphasizing the necessity of guidelines for researchers and practitioners to promote responsible use while safeguarding individual rights and societal norms.

%% file: sections/conclusion_arxiv.tex
\section{Conclusion}

In this paper, we propose OpenHumanVid, a large-scale, high-quality video dataset designed to address the lack of human-centric video data. Our dataset includes a variety of motion conditions, which can facilitate a wide range of human-centric tasks. By pretraining a simple extension of existing models on OpenHumanVid, we achieve improvements in human video generation performance. We believe our work will provide valuable resources for advancing the field of human video generation and support future research in this area.

\vspace{+2mm}

%% file: PaperForReview.bbl
\begin{thebibliography}{10}\itemsep=-1pt

\bibitem{baek2019character}
Youngmin Baek, Bado Lee, Dongyoon Han, Sangdoo Yun, and Hwalsuk Lee.
\newblock Character region awareness for text detection.
\newblock In {\em Proceedings of the IEEE Conference on Computer Vision and Pattern Recognition}, pages 9365--9374, 2019.

\bibitem{bain2021frozen}
Max Bain, Arsha Nagrani, G{\"u}l Varol, and Andrew Zisserman.
\newblock Frozen in time: A joint video and image encoder for end-to-end retrieval.
\newblock In {\em Proceedings of the IEEE/CVF international conference on computer vision}, pages 1728--1738, 2021.

\bibitem{bao2024vidu}
Fan Bao, Chendong Xiang, Gang Yue, Guande He, Hongzhou Zhu, Kaiwen Zheng, Min Zhao, Shilong Liu, Yaole Wang, and Jun Zhu.
\newblock Vidu: a highly consistent, dynamic and skilled text-to-video generator with diffusion models.
\newblock {\em arXiv preprint arXiv:2405.04233}, 2024.

\bibitem{black2023bedlam}
Michael~J Black, Priyanka Patel, Joachim Tesch, and Jinlong Yang.
\newblock Bedlam: A synthetic dataset of bodies exhibiting detailed lifelike animated motion.
\newblock In {\em Proceedings of the IEEE/CVF Conference on Computer Vision and Pattern Recognition}, pages 8726--8737, 2023.

\bibitem{blattmann2023stable}
Andreas Blattmann, Tim Dockhorn, Sumith Kulal, Daniel Mendelevitch, Maciej Kilian, Dominik Lorenz, Yam Levi, Zion English, Vikram Voleti, Adam Letts, Varun Jampani, and Robin Rombach.
\newblock Stable video diffusion: Scaling latent video diffusion models to large datasets.
\newblock {\em arXiv preprint arXiv:2311.15127}, 2023.

\bibitem{blattmann2023align}
Andreas Blattmann, Robin Rombach, Huan Ling, Tim Dockhorn, Seung~Wook Kim, Sanja Fidler, and Karsten Kreis.
\newblock Align your latents: High-resolution video synthesis with latent diffusion models.
\newblock In {\em Proceedings of the IEEE/CVF Conference on Computer Vision and Pattern Recognition}, 2023.

\bibitem{caba2015activitynet}
Fabian Caba~Heilbron, Victor Escorcia, Bernard Ghanem, and Juan Carlos~Niebles.
\newblock Activitynet: A large-scale video benchmark for human activity understanding.
\newblock In {\em Proceedings of the ieee conference on computer vision and pattern recognition}, pages 961--970, 2015.

\bibitem{camgoz2018neural}
Necati~Cihan Camgoz, Simon Hadfield, Oscar Koller, Hermann Ney, and Richard Bowden.
\newblock Neural sign language translation.
\newblock In {\em Proceedings of the IEEE conference on computer vision and pattern recognition}, pages 7784--7793, 2018.

\bibitem{caron2021emerging}
Mathilde Caron, Hugo Touvron, Ishan Misra, Herv\'e J\'egou, Julien Mairal, Piotr Bojanowski, and Armand Joulin.
\newblock Emerging properties in self-supervised vision transformers.
\newblock In {\em Proceedings of the International Conference on Computer Vision (ICCV)}, 2021.

\bibitem{PySceneDetect}
Brandon Castellano.
\newblock Pyscenedetect.
\newblock \url{https://github.com/Breakthrough/PySceneDetect}.

\bibitem{chang2023magicdance}
Di Chang, Yichun Shi, Quankai Gao, Jessica Fu, Hongyi Xu, Guoxian Song, Qing Yan, Xiao Yang, and Mohammad Soleymani.
\newblock Magicdance: Realistic human dance video generation with motions \& facial expressions transfer.
\newblock {\em arXiv preprint arXiv:2311.12052}, 2023.

\bibitem{chang2023magicpose}
Di Chang, Yichun Shi, Quankai Gao, Hongyi Xu, Jessica Fu, Guoxian Song, Qing Yan, Yizhe Zhu, Xiao Yang, and Mohammad Soleymani.
\newblock Magicpose: Realistic human poses and facial expressions retargeting with identity-aware diffusion.
\newblock In {\em Forty-first International Conference on Machine Learning}, 2023.

\bibitem{chen2024multi}
Hong Chen, Xin Wang, Yuwei Zhou, Bin Huang, Yipeng Zhang, Wei Feng, Houlun Chen, Zeyang Zhang, Siao Tang, and Wenwu Zhu.
\newblock Multi-modal generative ai: Multi-modal llm, diffusion and beyond.
\newblock {\em arXiv preprint arXiv:2409.14993}, 2024.

\bibitem{chen2024panda}
Tsai-Shien Chen, Aliaksandr Siarohin, Willi Menapace, Ekaterina Deyneka, Hsiang-wei Chao, Byung~Eun Jeon, Yuwei Fang, Hsin-Ying Lee, Jian Ren, Ming-Hsuan Yang, et~al.
\newblock Panda-70m: Captioning 70m videos with multiple cross-modality teachers.
\newblock In {\em Proceedings of the IEEE/CVF Conference on Computer Vision and Pattern Recognition}, pages 13320--13331, 2024.

\bibitem{corona2024vlogger}
Enric Corona, Andrei Zanfir, Eduard~Gabriel Bazavan, Nikos Kolotouros, Thiemo Alldieck, and Cristian Sminchisescu.
\newblock Vlogger: Multimodal diffusion for embodied avatar synthesis.
\newblock {\em arXiv preprint arXiv:2403.08764}, 2024.

\bibitem{cui2024hallo2}
Jiahao Cui, Hui Li, Yao Yao, Hao Zhu, Hanlin Shang, Kaihui Cheng, Hang Zhou, Siyu Zhu, and Jingdong Wang.
\newblock Hallo2: Long-duration and high-resolution audio-driven portrait image animation.
\newblock {\em arXiv preprint arXiv:2410.07718}, 2024.

\bibitem{Deng2020CVPR}
Jiankang Deng, Jia Guo, Evangelos Ververas, Irene Kotsia, and Stefanos Zafeiriou.
\newblock Retinaface: Single-shot multi-level face localisation in the wild.
\newblock In {\em CVPR}, 2020.

\bibitem{dubey2024LlaMA}
Abhimanyu Dubey, Abhinav Jauhri, Abhinav Pandey, Abhishek Kadian, Ahmad Al-Dahle, Aiesha Letman, Akhil Mathur, Alan Schelten, Amy Yang, Angela Fan, et~al.
\newblock The llama 3 herd of models.
\newblock {\em arXiv preprint arXiv:2407.21783}, 2024.

\bibitem{he2023gaia}
Tianyu He, Junliang Guo, Runyi Yu, Yuchi Wang, Jialiang Zhu, Kaikai An, Leyi Li, Xu Tan, Chunyu Wang, Han Hu, et~al.
\newblock Gaia: Zero-shot talking avatar generation.
\newblock {\em arXiv preprint arXiv:2311.15230}, 2023.

\bibitem{ho2022imagen}
Jonathan Ho, William Chan, Chitwan Saharia, Jay Whang, Ruiqi Gao, Alexey Gritsenko, Diederik~P Kingma, Ben Poole, Mohammad Norouzi, David~J Fleet, et~al.
\newblock Imagen video: High definition video generation with diffusion models.
\newblock {\em arXiv preprint arXiv:2210.02303}, 2022.

\bibitem{ho2021classifier}
Jonathan Ho and Tim Salimans.
\newblock Classifier-free diffusion guidance.
\newblock In {\em NeurIPS 2021 Workshop on Deep Generative Models and Downstream Applications}, 2021.

\bibitem{ho2022video}
Jonathan Ho, Tim Salimans, Alexey Gritsenko, William Chan, Mohammad Norouzi, and David~J Fleet.
\newblock Video diffusion models.
\newblock {\em arXiv:2204.03458}, 2022.

\bibitem{hong2024cogvlm2}
Wenyi Hong, Weihan Wang, Ming Ding, Wenmeng Yu, Qingsong Lv, Yan Wang, Yean Cheng, Shiyu Huang, Junhui Ji, Zhao Xue, et~al.
\newblock Cogvlm2: Visual language models for image and video understanding.
\newblock {\em arXiv preprint arXiv:2408.16500}, 2024.

\bibitem{hu2021lora}
Edward~J Hu, Phillip Wallis, Zeyuan Allen-Zhu, Yuanzhi Li, Shean Wang, Lu Wang, Weizhu Chen, et~al.
\newblock Lora: Low-rank adaptation of large language models.
\newblock In {\em International Conference on Learning Representations (ICLR)}, 2021.

\bibitem{hu2024animate}
Li Hu.
\newblock Animate anyone: Consistent and controllable image-to-video synthesis for character animation.
\newblock In {\em Proceedings of the IEEE/CVF Conference on Computer Vision and Pattern Recognition}, pages 8153--8163, 2024.

\bibitem{huang2024vbench}
Ziqi Huang, Yinan He, Jiashuo Yu, Fan Zhang, Chenyang Si, Yuming Jiang, Yuanhan Zhang, Tianxing Wu, Qingyang Jin, Nattapol Chanpaisit, et~al.
\newblock Vbench: Comprehensive benchmark suite for video generative models.
\newblock In {\em Proceedings of the IEEE/CVF Conference on Computer Vision and Pattern Recognition}, pages 21807--21818, 2024.

\bibitem{jafarian2021learning}
Yasamin Jafarian and Hyun~Soo Park.
\newblock Learning high fidelity depths of dressed humans by watching social media dance videos.
\newblock In {\em Proceedings of the IEEE/CVF Conference on Computer Vision and Pattern Recognition}, pages 12753--12762, 2021.

\bibitem{jiang2023text2performer}
Yuming Jiang, Shuai Yang, Tong~Liang Koh, Wayne Wu, Chen~Change Loy, and Ziwei Liu.
\newblock Text2performer: Text-driven human video generation.
\newblock In {\em Proceedings of the IEEE/CVF International Conference on Computer Vision}, pages 22747--22757, 2023.

\bibitem{ju2023human}
Xuan Ju, Ailing Zeng, Jianan Wang, Qiang Xu, and Lei Zhang.
\newblock Human-art: A versatile human-centric dataset bridging natural and artificial scenes.
\newblock In {\em Proceedings of the IEEE/CVF Conference on Computer Vision and Pattern Recognition}, pages 618--629, 2023.

\bibitem{kingma2013auto}
Diederik~P Kingma.
\newblock Auto-encoding variational bayes.
\newblock {\em arXiv preprint arXiv:1312.6114}, 2013.

\bibitem{kingma2022autoencoding}
Diederik~P Kingma and Max Welling.
\newblock Auto-encoding variational bayes, 2022.

\bibitem{kondratyuk2023videopoet}
Dan Kondratyuk, Lijun Yu, Xiuye Gu, Jos{\'e} Lezama, Jonathan Huang, Grant Schindler, Rachel Hornung, Vighnesh Birodkar, Jimmy Yan, Ming-Chang Chiu, et~al.
\newblock Videopoet: A large language model for zero-shot video generation.
\newblock {\em arXiv preprint arXiv:2312.14125}, 2023.

\bibitem{li2023blip}
Junnan Li, Dongxu Li, Silvio Savarese, and Steven Hoi.
\newblock Blip-2: Bootstrapping language-image pre-training with frozen image encoders and large language models.
\newblock In {\em International conference on machine learning}, pages 19730--19742. PMLR, 2023.

\bibitem{liu2024sora}
Yixin Liu, Kai Zhang, Yuan Li, Zhiling Yan, Chujie Gao, Ruoxi Chen, Zhengqing Yuan, Yue Huang, Hanchi Sun, Jianfeng Gao, et~al.
\newblock Sora: A review on background, technology, limitations, and opportunities of large vision models.
\newblock {\em arXiv preprint arXiv:2402.17177}, 2024.

\bibitem{miech2019howto100m}
Antoine Miech, Dimitri Zhukov, Jean-Baptiste Alayrac, Makarand Tapaswi, Ivan Laptev, and Josef Sivic.
\newblock Howto100m: Learning a text-video embedding by watching hundred million narrated video clips.
\newblock In {\em Proceedings of the IEEE/CVF international conference on computer vision}, pages 2630--2640, 2019.

\bibitem{nan2024openvid}
Kepan Nan, Rui Xie, Penghao Zhou, Tiehan Fan, Zhenheng Yang, Zhijie Chen, Xiang Li, Jian Yang, and Ying Tai.
\newblock Openvid-1m: A large-scale high-quality dataset for text-to-video generation.
\newblock {\em arXiv preprint arXiv:2407.02371}, 2024.

\bibitem{narvekar2011no}
Niranjan~D Narvekar and Lina~J Karam.
\newblock A no-reference image blur metric based on the cumulative probability of blur detection (cpbd).
\newblock {\em IEEE Transactions on Image Processing}, 20(9):2678--2683, 2011.

\bibitem{peebles2023scalable}
William Peebles and Saining Xie.
\newblock Scalable diffusion models with transformers.
\newblock In {\em Proceedings of the IEEE/CVF International Conference on Computer Vision}, pages 4195--4205, 2023.

\bibitem{polyak2024movie}
Adam Polyak, Amit Zohar, Andrew Brown, Andros Tjandra, Animesh Sinha, Ann Lee, Apoorv Vyas, Bowen Shi, Chih-Yao Ma, Ching-Yao Chuang, et~al.
\newblock Movie gen: A cast of media foundation models.
\newblock {\em arXiv preprint arXiv:2410.13720}, 2024.

\bibitem{raffel2020exploring}
Colin Raffel, Noam Shazeer, Adam Roberts, Katherine Lee, Sharan Narang, Michael Matena, Yanqi Zhou, Wei Li, and Peter~J Liu.
\newblock Exploring the limits of transfer learning with a unified text-to-text transformer.
\newblock {\em Journal of machine learning research}, 21(140):1--67, 2020.

\bibitem{raina2022syncnet}
Akshay Raina and Vipul Arora.
\newblock Syncnet: Using causal convolutions and correlating objective for time delay estimation in audio signals.
\newblock {\em arXiv preprint arXiv:2203.14639}, 2022.

\bibitem{rombach2022high}
Robin Rombach, Andreas Blattmann, Dominik Lorenz, Patrick Esser, and Bj{\"o}rn Ommer.
\newblock High-resolution image synthesis with latent diffusion models.
\newblock In {\em Proceedings of the IEEE/CVF conference on computer vision and pattern recognition}, pages 10684--10695, 2022.

\bibitem{sadoughi2015msp}
Najmeh Sadoughi, Yang Liu, and Carlos Busso.
\newblock Msp-avatar corpus: Motion capture recordings to study the role of discourse functions in the design of intelligent virtual agents.
\newblock In {\em 2015 11th IEEE International Conference and Workshops on Automatic Face and Gesture Recognition (FG)}, volume~7, pages 1--6. IEEE, 2015.

\bibitem{aesthetic_predictor}
Christoph Schuhmann.
\newblock aesthetic predictor.
\newblock \url{https://github.com/christophschuhmann/improved-aesthetic-predictor}.

\bibitem{shahroudy2016ntu}
Amir Shahroudy, Jun Liu, Tian-Tsong Ng, and Gang Wang.
\newblock Ntu rgb+ d: A large scale dataset for 3d human activity analysis.
\newblock In {\em Proceedings of the IEEE conference on computer vision and pattern recognition}, pages 1010--1019, 2016.

\bibitem{soomro2012ucf101}
K Soomro.
\newblock Ucf101: A dataset of 101 human actions classes from videos in the wild.
\newblock {\em arXiv preprint arXiv:1212.0402}, 2012.

\bibitem{su2024roformer}
Jianlin Su, Murtadha Ahmed, Yu Lu, Shengfeng Pan, Wen Bo, and Yunfeng Liu.
\newblock Roformer: Enhanced transformer with rotary position embedding.
\newblock {\em Neurocomputing}, 568:127063, 2024.

\bibitem{vaswani2017attention}
A Vaswani.
\newblock Attention is all you need.
\newblock {\em Advances in Neural Information Processing Systems}, 2017.

\bibitem{villegas2022phenaki}
Ruben Villegas, Mohammad Babaeizadeh, Pieter-Jan Kindermans, Hernan Moraldo, Han Zhang, Mohammad~Taghi Saffar, Santiago Castro, Julius Kunze, and Dumitru Erhan.
\newblock Phenaki: Variable length video generation from open domain textual descriptions.
\newblock In {\em International Conference on Learning Representations}, 2022.

\bibitem{wang2023modelscope}
Jiuniu Wang, Hangjie Yuan, Dayou Chen, Yingya Zhang, Xiang Wang, and Shiwei Zhang.
\newblock Modelscope text-to-video technical report.
\newblock {\em arXiv preprint arXiv:2308.06571}, 2023.

\bibitem{wang2024koala}
Qiuheng Wang, Yukai Shi, Jiarong Ou, Rui Chen, Ke Lin, Jiahao Wang, Boyuan Jiang, Haotian Yang, Mingwu Zheng, Xin Tao, et~al.
\newblock Koala-36m: A large-scale video dataset improving consistency between fine-grained conditions and video content.
\newblock {\em arXiv preprint arXiv:2410.08260}, 2024.

\bibitem{wang2024disco}
Tan Wang, Linjie Li, Kevin Lin, Yuanhao Zhai, Chung-Ching Lin, Zhengyuan Yang, Hanwang Zhang, Zicheng Liu, and Lijuan Wang.
\newblock Disco: Disentangled control for realistic human dance generation.
\newblock In {\em Proceedings of the IEEE/CVF Conference on Computer Vision and Pattern Recognition}, pages 9326--9336, 2024.

\bibitem{wang2024dance}
Xuanchen Wang, Heng Wang, Dongnan Liu, and Weidong Cai.
\newblock Dance any beat: Blending beats with visuals in dance video generation.
\newblock {\em arXiv preprint arXiv:2405.09266}, 2024.

\bibitem{wang2023internvid}
Yi Wang, Yinan He, Yizhuo Li, Kunchang Li, Jiashuo Yu, Xin Ma, Xinhao Li, Guo Chen, Xinyuan Chen, Yaohui Wang, et~al.
\newblock Internvid: A large-scale video-text dataset for multimodal understanding and generation.
\newblock {\em arXiv preprint arXiv:2307.06942}, 2023.

\bibitem{wang2024humanvid}
Zhenzhi Wang, Yixuan Li, Yanhong Zeng, Youqing Fang, Yuwei Guo, Wenran Liu, Jing Tan, Kai Chen, Tianfan Xue, Bo Dai, et~al.
\newblock Humanvid: Demystifying training data for camera-controllable human image animation.
\newblock {\em arXiv preprint arXiv:2407.17438}, 2024.

\bibitem{weng2021convolutional}
W Weng and X~Zhu INet.
\newblock Convolutional networks for biomedical image segmentation., 2021, 9.
\newblock {\em DOI: https://doi. org/10.1109/ACCESS}, pages 16591--16603, 2021.

\bibitem{wu2023exploring}
Haoning Wu, Erli Zhang, Liang Liao, Chaofeng Chen, Jingwen Hou, Annan Wang, Wenxiu Sun, Qiong Yan, and Weisi Lin.
\newblock Exploring video quality assessment on user generated contents from aesthetic and technical perspectives.
\newblock In {\em Proceedings of the IEEE/CVF International Conference on Computer Vision}, pages 20144--20154, 2023.

\bibitem{xu2024hallo}
Mingwang Xu, Hui Li, Qingkun Su, Hanlin Shang, Liwei Zhang, Ce Liu, Jingdong Wang, Yao Yao, and Siyu zhu.
\newblock Hallo: Hierarchical audio-driven visual synthesis for portrait image animation, 2024.

\bibitem{xu2024vasa}
Sicheng Xu, Guojun Chen, Yu-Xiao Guo, Jiaolong Yang, Chong Li, Zhenyu Zang, Yizhong Zhang, Xin Tong, and Baining Guo.
\newblock Vasa-1: Lifelike audio-driven talking faces generated in real time.
\newblock {\em arXiv preprint arXiv:2404.10667}, 2024.

\bibitem{xue2022advancing}
Hongwei Xue, Tiankai Hang, Yanhong Zeng, Yuchong Sun, Bei Liu, Huan Yang, Jianlong Fu, and Baining Guo.
\newblock Advancing high-resolution video-language representation with large-scale video transcriptions.
\newblock In {\em Proceedings of the IEEE/CVF Conference on Computer Vision and Pattern Recognition}, pages 5036--5045, 2022.

\bibitem{unimatch}
Lihe Yang, Lei Qi, Litong Feng, Wayne Zhang, and Yinghuan Shi.
\newblock Revisiting weak-to-strong consistency in semi-supervised semantic segmentation.
\newblock In {\em CVPR}, 2023.

\bibitem{yang2023qpgesture}
Sicheng Yang, Zhiyong Wu, Minglei Li, Zhensong Zhang, Lei Hao, Weihong Bao, and Haolin Zhuang.
\newblock Qpgesture: Quantization-based and phase-guided motion matching for natural speech-driven gesture generation.
\newblock In {\em Proceedings of the IEEE/CVF Conference on Computer Vision and Pattern Recognition}, pages 2321--2330, 2023.

\bibitem{yang2024cogvideox}
Zhuoyi Yang, Jiayan Teng, Wendi Zheng, Ming Ding, Shiyu Huang, Jiazheng Xu, Yuanming Yang, Wenyi Hong, Xiaohan Zhang, Guanyu Feng, et~al.
\newblock Cogvideox: Text-to-video diffusion models with an expert transformer.
\newblock {\em arXiv preprint arXiv:2408.06072}, 2024.

\bibitem{yang2023effective}
Zhendong Yang, Ailing Zeng, Chun Yuan, and Yu Li.
\newblock Effective whole-body pose estimation with two-stages distillation.
\newblock In {\em Proceedings of the IEEE/CVF International Conference on Computer Vision}, 2023.

\bibitem{yao2024minicpm}
Yuan Yao, Tianyu Yu, Ao Zhang, Chongyi Wang, Junbo Cui, Hongji Zhu, Tianchi Cai, Haoyu Li, Weilin Zhao, Zhihui He, et~al.
\newblock Minicpm-v: A gpt-4v level mllm on your phone.
\newblock {\em arXiv preprint arXiv:2408.01800}, 2024.

\bibitem{zhang2024tora}
Zhenghao Zhang, Junchao Liao, Menghao Li, Long Qin, and Weizhi Wang.
\newblock Tora: Trajectory-oriented diffusion transformer for video generation.
\newblock {\em arXiv preprint arXiv:2407.21705}, 2024.

\bibitem{opensora}
Zangwei Zheng, Xiangyu Peng, Tianji Yang, Chenhui Shen, Shenggui Li, Hongxin Liu, Yukun Zhou, Tianyi Li, and Yang You.
\newblock Open-sora: Democratizing efficient video production for all, March 2024.

\bibitem{zhu2023taming}
Lingting Zhu, Xian Liu, Xuanyu Liu, Rui Qian, Ziwei Liu, and Lequan Yu.
\newblock Taming diffusion models for audio-driven co-speech gesture generation.
\newblock In {\em Proceedings of the IEEE/CVF Conference on Computer Vision and Pattern Recognition}, pages 10544--10553, 2023.

\bibitem{zhu2024champ}
Shenhao Zhu, Junming~Leo Chen, Zuozhuo Dai, Yinghui Xu, Xun Cao, Yao Yao, Hao Zhu, , and Siyu Zhu.
\newblock Champ: Controllable and consistent human image animation with 3d parametric guidance.
\newblock {\em arXiv preprint arXiv:2403.14781}, 2024.

\end{thebibliography}
